\documentclass[10pt,twocolumn,letterpaper]{article}

\usepackage[pagenumbers]{cvpr} 

\usepackage{graphicx}
\usepackage{amsmath}
\usepackage{amssymb}
\usepackage{booktabs}

\usepackage{color}

\usepackage{multirow}
\usepackage{makecell}

\makeatletter
\newcommand*{\overrightharpoonup}{\mathpalette{\overarrow@\rightharpoonupfill@}}
\newcommand*{\rightharpoonupfill@}{\arrowfill@\relbar\relbar\rightharpoonup}
\makeatother

\usepackage[pagebackref,breaklinks,colorlinks]{hyperref}

\usepackage[capitalize]{cleveref}
\crefname{section}{Sec.}{Secs.}
\Crefname{section}{Section}{Sections}
\Crefname{table}{Table}{Tables}
\crefname{table}{Tab.}{Tabs.}

\def\cvprPaperID{5667} 
\def\confName{CVPR}
\def\confYear{2022}

\begin{document}

\title{Deeply Interleaved Two-Stream Encoder for Referring Video Segmentation}

\author{Guang Feng, Lihe Zhang\footnotemark[2], Zhiwei Hu, Huchuan Lu\\
\small $^1$School of Information and Communication Engineering, Dalian University of Technology \\
{\tt\small fengguang.gg@gmail.com, hzw950822@mail.dlut.edu.cn, }\\
{\tt\small \{zhanglihe, lhchuan\}@dlut.edu.cn}\\
}
\maketitle
\thispagestyle{empty}
\renewcommand{\thefootnote}{\fnsymbol{footnote}}
\footnotetext[2]{Corresponding Author}

\begin{abstract}
   Referring video segmentation aims to segment the corresponding video object described by the language expression. To address this task, we first design a two-stream encoder to extract CNN-based visual features and transformer-based linguistic features hierarchically, and a vision-language mutual guidance (VLMG) module is inserted into the encoder multiple times to promote the hierarchical and progressive fusion of multi-modal features. Compared with the existing multi-modal fusion methods, this two-stream encoder takes into account the multi-granularity linguistic context, and realizes the deep interleaving between modalities with the help of VLGM. In order to promote the temporal alignment between frames, we further propose a language-guided multi-scale dynamic filtering (LMDF) module to strengthen the temporal coherence, which uses the language-guided spatial-temporal features to generate a set of position-specific dynamic filters to more flexibly and effectively update the feature of current frame. Extensive experiments on four datasets verify the effectiveness of the proposed model.
\end{abstract}

\section{Introduction}
\label{sec:intro}

This paper aims to address a task called referring video segmentation. Given a video clip and a referring expression, the goal of this task is to segment the corresponding entity (object) in the video sequence according to the description of query language. Traditional semi-supervised video object segmentation uses the ground-truth mask of the first frame (or a few frames) as the guide to segment the corresponding foreground object(s) in the remaining video sequence. Similarly, referring video segmentation considers language expression as interactive information to guide the segmentation of video object. Compared with the ground-truth mask, the language labeling is more flexible and economy. This task can be applied to language-driven video editing, human-robot interaction, etc.

\begin{figure}[t]
\vspace{0mm}
\includegraphics[width=1.0\linewidth]{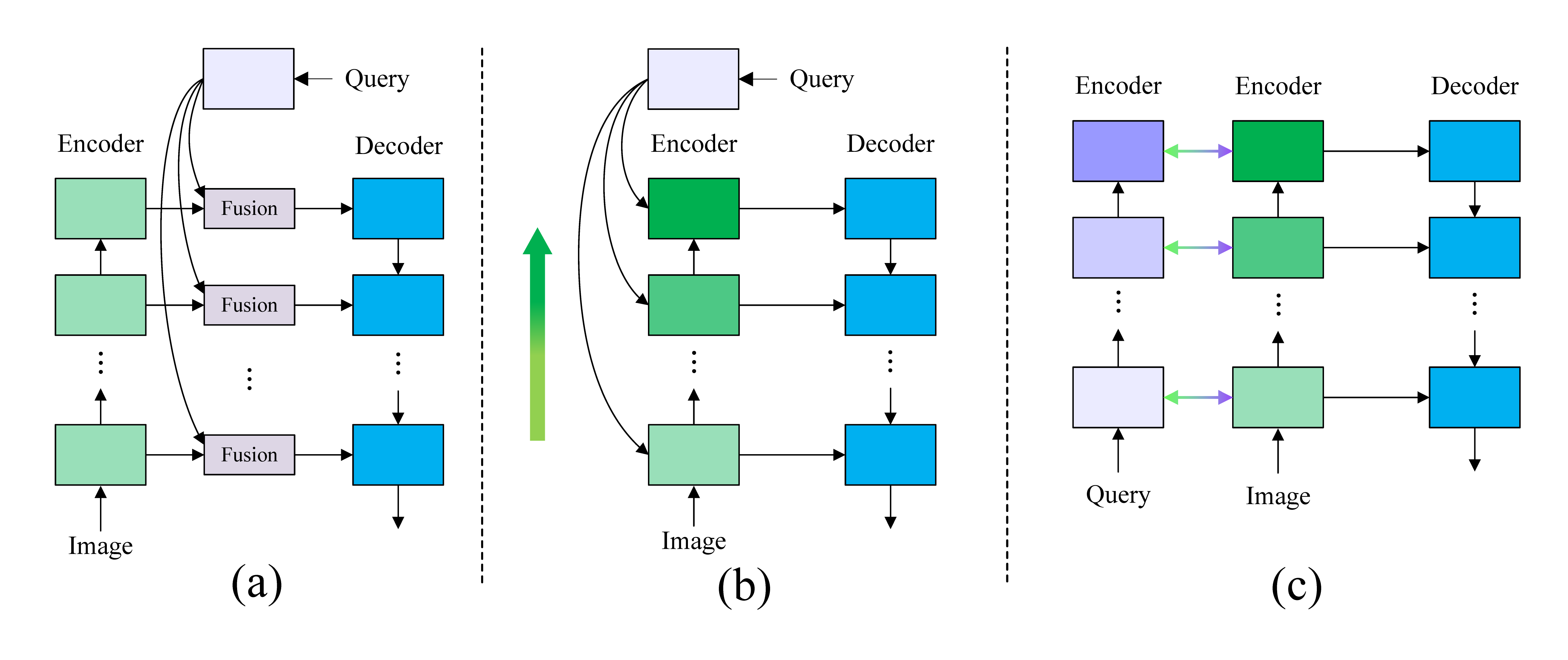}
\ \\
{\begin{center}
\vspace{-12mm}
\caption{\small{Three multi-modal fusion mechanisms. (a) Decoder fusion. (b) Unidirectional encoder fusion. (c) Bidirectional encoder fusion, which realizes the hierarchical cross-modal interaction between vision and language.
}}
\label{fig:introduction}
\end{center}
}
\vspace{-3mm}
\end{figure}

Referring video segmentation contains two crucial issues. They are how to achieve the modality consistency between linguistic and visual information and how to leverage the temporal coherency between frames. For the first issue, the referring image/video segmentation methods that have emerged in recent years adopt straightforward concatenation-convolution~\cite{hu2016segmentation}, recurrent LSTM~\cite{liu2017recurrent}, convolution LSTM~\cite{li2018referring}, dynamic filters~\cite{margffoy2018dynamic,gavrilyuk2018actor,wang2020context}, cross-modal attention ~\cite{shi2018key,ye2019cross,wang2019asymmetric,hu2020bi,huang2020referring,hui2020linguistic,seo2020urvos,feng2021encoder,hui2021collaborative,ding2021vision} to aggregate linguistic and visual features.
According to the position of multi-modal fusion in the network, they can be divided into two categories: decoder fusion (Fig.~\ref{fig:introduction} (a)) and encoder fusion (Fig.~\ref{fig:introduction} (b)).
The former uses linguistic feature to separately guide visual features of different levels in the decoder for cross-modal embedding. The latter realizes the progressive guidance of language to the multi-modal features in the encoder~\cite{feng2021encoder}.
However, during the process of linguistic feature extraction, both of them ignore the semantic hierarchy of language information similar to the multi-level property embodied by visual features, which may weaken the consistency of the multiple-granularity cross-modal alignment.

Regarding the second issue, most of the existing methods~\cite{gavrilyuk2018actor,wang2019asymmetric,wang2020context,mcintosh2020visual} directly employ a I3D encoder~\cite{carreira2017quo} to extract visual features for each frame, which implicitly characterizes the temporal relevance among different frames. The visual features are then handled the same way as image referring segmentation methods do. These video based methods actually focus on the design of cross-modal fusion while neglecting the inter-frame relationship modeling.
Moreover, they do not utilize the language to guidance inter-frame feature interaction. These flaws inevitably weaken the temporal coherence among the features of frames.
In~\cite{hui2021collaborative}, the language-guided channel attention is designed to re-weight and combine the features of current frame and reference frame, in which their attention weights are query-specific and are shared across the whole video. This design of temporal modeling does not consider the inter-frame adaptability and the intra-frame inhomogeneity well.

To address the above-mentioned problems, we introduce the vision-language interleaved encoder and the language-guided dynamic filtering mechanism. Firstly, from the perspective
of vision-language fusion, we adopt CNN encoder and transformer encoder to extract visual and linguistic features, respectively. Meanwhile, a vision-language mutual guidance (VLMG) module is repeatedly embedded between the two encoding streams to gradually transfer hierarchical cross-modal information to each other. Thus, the cross-modal alignment can be achieved at multiple levels of granularity. Secondly, for the temporal modeling, we propose a language-guided multi-scale dynamic filtering (LMDF) module, which exploits the spatial-temporal features to learn a set of position-specific and frame-specific dynamic filters under the guidance of language, and then utilize these filters to update the multi-level features of current frame. This module can promote the network to steadily and accurately capture the visual clues referred by the language query along the video.

Our main contributions are as follows:
\vspace{0mm}
\begin{itemize}
\vspace{-0mm}
\item We propose a vision-language interleaved two-stream encoder, which leverages the vision-language mutual guidance module to effectively realize the multi-level information interaction between vision and language encoding steams. The interleaved encoder can produce a compact multi-modal representation.
\vspace{-0mm}
\item We design a language-guided multi-scale dynamic filtering module, which utilizes the language-guided spatial-temporal context to generate region-aware dynamic filters to update the features of current frame, thereby ensuring the temporal coherence of multi-modal feature learning.
\vspace{-0mm}
\item Extensive evaluations on four referring video segmentation datasets (A2D, J-HMDB, Ref-DAVIS2017 and Ref-Youtube-VOS) demonstrate that our method surpasses the previous state-of-the-art approaches.
\end{itemize}


\section{Related Work}
\label{sec:related}

\begin{figure*}[t]
\centering
\includegraphics[width=1.0\linewidth]{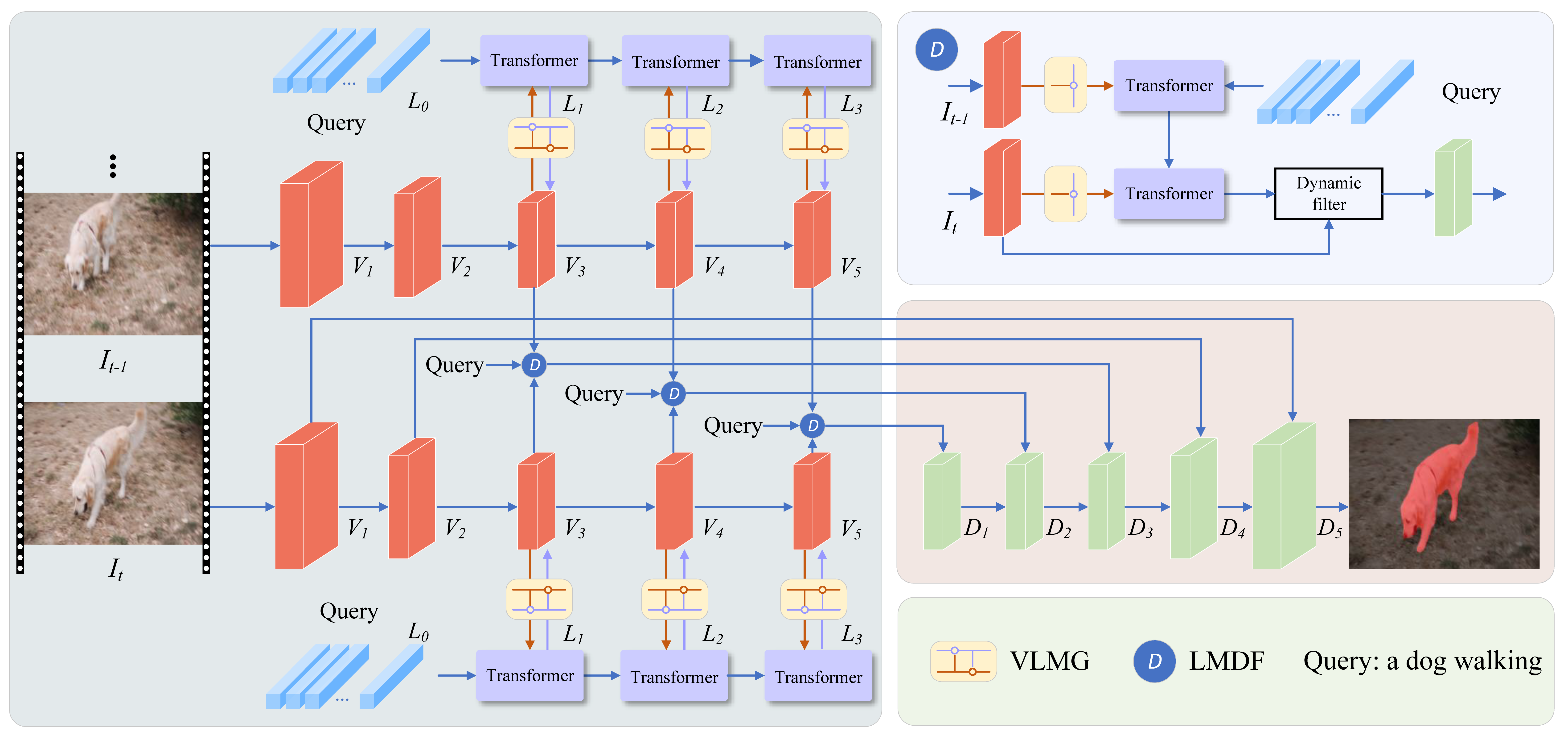}
\ \\
{\begin{center}
\vspace{-3mm}
\caption{\small{The overall architecture of our model. It mainly consists of the transformer-based linguistic encoder, CNN-based visual encoder, vision-language mutual guidance (VLMG) module, language-guided multi-scale dynamic filtering (LMDF) module. $I_t$ denotes the current frame. $I_{t-1}$ denotes the reference frame.}}
\label{fig:structure}
\end{center}
}
\vspace{-2mm}
\end{figure*}
\subsection{Vision-Language Interaction}
%
Early method~\cite{hu2016segmentation} utilizes concatenation and convolution to fuse linguistic and visual features. Subsequently, Liu \emph{et al.}~\cite{liu2017recurrent} and Li \emph{et al.}~\cite{li2018referring} adopt recurrent LSTM and convolutional LSTM to gradually combine the concatenated multi-modal features in the decoder, respectively. These methods directly concatenate visual and linguistic features, which dose not explicitly consider the relationship between each pixel and each word.

To solve this problem, some methods exploit attention mechanism to model the cross-modal relationship. For example, Shi \emph{et al.}~\cite{shi2018key} use the visual features as a guidance to learn the keywords corresponding to each pixel, which can suppress the noise in the query expression. Ye \emph{et al.}~\cite{ye2019cross} and Seo \emph{et al.}~\cite{seo2020urvos} employ non-local module to model the relationship between the pixel-word mixed features. Wang \emph{et al.}~\cite{wang2019asymmetric} propose an asymmetric cross-attention module, which utilizes two affinity matrices to update visual and linguistic features, respectively. Hu \emph{et al.}~\cite{hu2020bi} build a bidirectional relationship decoder to realize the mutual guidance between language and vision. Hui \emph{et al.}~\cite{hui2020linguistic} rely on the dependency parsing tree to re-weight the edges of the word graph, thereby suppressing the useless edges in the initial fully connected graph. Huang \emph{et al.}~\cite{huang2020referring} first perceive all entities in the image according to the category and appearance information and then employ the relation words to model the relationships among all entities.  Ding \emph{et al.}~\cite{ding2021vision} design a transformer-based query generation module at the decoding end to understand the vision-language context.

Some other methods employ the dynamic filters generated by the sentence to match the visual features, thereby strengthening the response of the language-related visual region. Gavrilyuk \emph{et al.}~\cite{gavrilyuk2018actor} directly apply a dynamic filter to weight all channels of feature map and sum them to yield the pixel-wise segmentation map.
Similarly, Margffoy-Tuay \emph{et al.}~\cite{margffoy2018dynamic} learn a set of dynamic filters for each word to re-weight the visual features. Wang \emph{et al.}~\cite{wang2020context} introduce deformable convolution~\cite{dai2017deformable} into the process of dynamic filtering. In addition. McIntosh \emph{et al.}~\cite{mcintosh2020visual} design a capsule network to perform vision-language coding.

A common feature of all the above methods is that they fuse visual and linguistic features at the decoding end of the network. The interaction between visual feature of each scale and language feature is isolated. Recently, the encoder fusion strategy~\cite{ningpolar,feng2021encoder,hui2021collaborative} is applied to referring segmentation task. It achieves the continuous guidance of language to multi-scale visual features. Regardless of the encoder fusion  or the decoder fusion, their language encoding processes do not consider the boosting effect of visual information of multiple levels, and ignore the semantic hierarchy of linguistic information. Integrating the same language features with visual features of different levels may weaken the consistency of cross-modal matching. Different from them, we propose a vision-language interleaved two-stream encoder,  which can extract hierarchical language information with the help of multi-level visual features and implement the bidirectional cross-modal interaction at different levels.

\subsection{Temporal Feature Modeling}
Modeling temporal information between frames is important for referring video segmentation. A straightforward idea is to use 3D convolution to capture temporal cues of the video. Therefore, some methods~\cite{gavrilyuk2018actor,wang2019asymmetric,wang2020context,mcintosh2020visual} directly employ the I3D network~\cite{carreira2017quo} to encode video sequences. In the testing phase, the fixed convolution parameters are not easily generalized to all video clips, which limits the capability of temporal modeling.
To solve this problem, Seo \emph{et al.}~\cite{seo2020urvos} design a non-local based memory attention to learn the inter-frame correlation. Hui \emph{et al.}~\cite{hui2021collaborative} consider the cross-modal property of  referring video segmentation task and utilize language feature to build channel attention, which is applied to the weighted fusion of frames. In this paper, we leverage the language-guided spatial-temporal information to learn a set of position-specific and frame-specific dynamic filters, which can  fully exploit the inter-modality and inter-frame information interaction to progressively combine the cross-modal aligned features at multiple scales for current frame.

\section{Method}
The overall structure of the proposed model is shown in Fig.~\ref{fig:structure}. Firstly, the video features from the CNN-based encoder and language features from the transformer-based encoder are deeply interweaved  through the vision-language mutual guidance (VLMG) module. In the pipeline of network, after the initial interweaving, the features flowed into the trailing end of the encoder and the whole decoder are already the multi-modal mixed features.  Next, we propose a language-guided multi-scale dynamic filtering (LMDF) module to fuse temporal information across frames. The multiple-scale spatial-temporal features output by LMDF and the visual features from the first two encoder blocks of the visual encoder are fed to the decoder for progressive fusion and upsampling, thereby generating the final segmentation result. Note that, for the reference frame ($I_{t-1}$) and the current frame ($I_{t}$), their encoder parameters are shared.
\subsection{Vision-Language Interleaved Encoder}
\begin{figure}[t]
\vspace{0mm}
\includegraphics[width=1.0\linewidth]{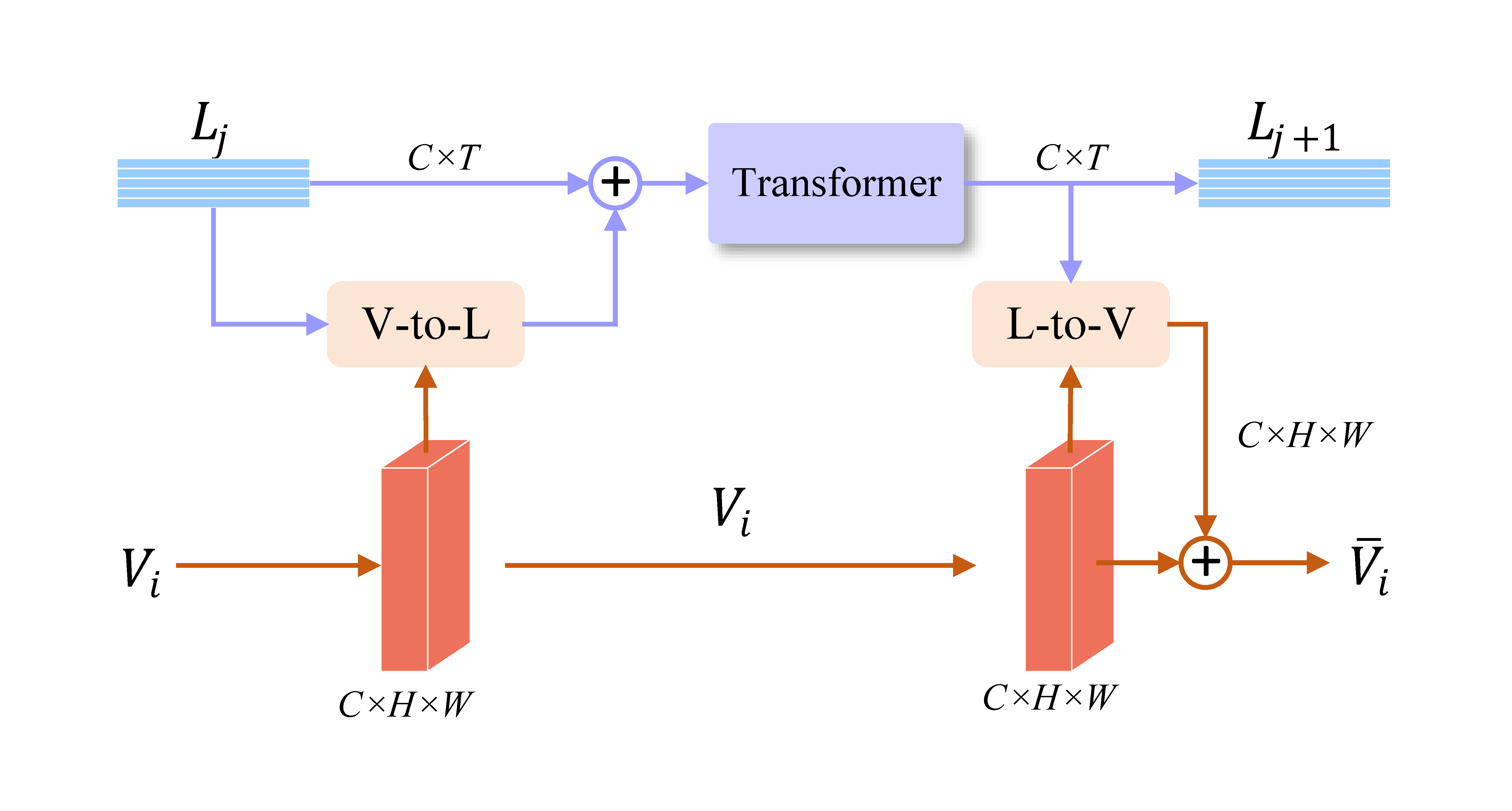}
\ \\
{\begin{center}
\vspace{-13mm}
\caption{\small{Vision-language mutual guidance (VLMG) module. V-to-L: Vision-to-language mapping. L-to-V: Language-to-vision mapping.
}}
\label{fig:VLMG}
\end{center}
}
\vspace{-2mm}
\end{figure}
We design a deeply interleaved two-stream encoder to achieve the mutual guidance and embedding between visual and linguistic features. It can not only encode the hierarchical semantic context of the sentence, but also gradually realize the mixing of multi-modal information from the feature extractors at different levels.

Specifically, we take ResNeSt~\cite{zhang2020resnest} as the visual encoder. It contains five feature blocks, and their features are defined as $\{ {\mathrm{V}_i} \} _{i=1}^5$. The query expression is denoted as $R=\{r_t\}^T_{t=1}$, where $t$ indicates the $t$-th word and $T$ is the total number of words. We utilize the Bert-embedding~\cite{devlin2018bert} to obtain the initial representation $\mathrm{L}_0=\{l_t^0\}^T_{t=1}$ of $R$. $\mathrm{L}_0$ is fed into three cascaded transformer blocks to encode the linguistic context $\{ {\mathrm{L}_j} \} _{j=1}^3$.

To establish the connection between the two encoders, we propose a vision-language mutual guidance (VLMG) module to interlacedly guide their feature extraction. For the convenience of description, we define the sizes of $\mathrm{V}_i$ and $\mathrm{L}_j$ as $C \times H \times W$ and $C \times T$, respectively. $H$, $W$ and $C$ denote the height, width and channel number, respectively.
The feature $\mathrm{V}_i$ is first reshaped into a matrix representation with size $C \times (HW)$. The vision-to-language affinity matrix between $\mathrm{V}_i$ and $\mathrm{L}_j$ can be calculated as follows:
\begin{equation}
\begin{split}
& {\mathrm{A}_{\scriptscriptstyle vl}}'= ({\mathrm{W}_v^1 \mathrm{V}_i)}^\top  ({ \mathrm{W}_l^1 \mathrm{L}_j) },\\
& {\mathrm{A}_{\scriptscriptstyle vl}}= {\rm softmax}({\mathrm{A}_{\scriptscriptstyle vl}}'),\\
\end{split} \label{v_to_l}
\end{equation}
where $\mathrm{W}_v^1$, $\mathrm{W}_l^1\in \mathbb{R}^{C_1\times C}$ are the learnable parameters. ${\mathrm{A}_{\scriptscriptstyle vl}} \in \mathbb{R}^{(HW)\times T}$ describes the similarity between each pixel of $\mathrm{V}_i$ and each word of $\mathrm{L}_j$. The softmax function is used to normalize the affinity matrix along the first dimension. We utilize ${\mathrm{A}_{\scriptscriptstyle vl}}$ to project the visual feature $\mathrm{V}_i$ to the language space as follows:
\begin{equation}
\begin{split}
& \widetilde{\mathrm{V}}_i = \mathrm{V}_i {\mathrm{A}_{\scriptscriptstyle vl}} .\\
\end{split}
\end{equation}

Next, the features $\widetilde{\mathrm{V}}_i \in \mathbb{R}^{C \times T}$ and $\mathrm{L}_j \in \mathbb{R}^{C \times T}$ are fed into the standard transformer structure to learn their co-embedding:
\begin{equation}
\begin{split}
& \widehat{\mathrm{L}}_j = \mathrm{MSA}(\mathrm{LN}(\widetilde{\mathrm{V}}_i + \mathrm{L}_j)) + (\widetilde{\mathrm{V}}_i + \mathrm{L}_j),\\
& \mathrm{L}_{j+1} = \mathrm{FFN}(\mathrm{LN}(\widehat{\mathrm{L}}_j)) + \widehat{\mathrm{L}}_j ,
\end{split} \label{transformer_block}
\end{equation}
where $\mathrm{MSA}$ denotes the multi-head self-attention module. $\mathrm{LN}$ represents the LayerNorm layer. FFN is the feed forward network. We further use $\mathrm{L}_{j+1} \in \mathbb{R}^{C \times T}$ and $\mathrm{V}_i$ to learn the affinity matrix of language-to-vision:
\begin{equation}
\begin{split}
& {\mathrm{A}_{\scriptscriptstyle lv}}'= ({\mathrm{W}_v^2 \mathrm{V}_i)}^\top  ({ \mathrm{W}_l^2 \mathrm{L}_{j+1}) },\\
& {\mathrm{A}_{\scriptscriptstyle lv}}= {\rm softmax}({{\mathrm{A}_{\scriptscriptstyle lv}}'}^\top),\\
\end{split} \label{l_to_v}
\end{equation}
where $\mathrm{W}_v^2$, $\mathrm{W}_l^2$ are the learned parameters. Similarly, the column-wise normalized affinity matrix ${\mathrm{A}_{\scriptscriptstyle lv}} \in \mathbb{R}^{T \times (HW)}$ is utilized to map the mixed multi-modal feature $\mathrm{L}_{j+1}$ to the vision space:
\begin{equation}
\begin{split}
& \widetilde{\mathrm{L}}_{j+1} = \mathrm{L}_{j+1} \mathrm{A}_{\scriptscriptstyle lv}.\\
\end{split}
\end{equation}

Finally, we adopt concatenation $Cat(\cdot,\cdot)$ and convolution $Conv$ to incorporate the feature $\widetilde{\mathrm{L}}_{j+1} \in \mathbb{R}^{C \times H \times W}$ into the visual encoder:
\begin{equation}
\begin{split}
&\overline{\mathrm{V}}_i = \mathrm{Norm}(\mathrm{V}_i) + \mathrm{Norm}(Conv(Cat(\mathrm{V}_i, \widetilde{\mathrm{L}}_{j+1}))) ,\\
\end{split}  \label{multi_out}
\end{equation}
where the $\mathrm{Norm}$ represents the L2-normalize, and its purpose is to normalize the feature maps to the same order of magnitude to avoid preference bias. The updated multi-modal feature $\overline{\mathrm{V}}_i$ is fed to the next convolution block of the visual encoder for further feature extraction.
Fig.~\ref{fig:VLMG} illustrates the detailed structure of VLMG.
By embedding VLMG between the visual CNN encoder and the linguistic transformer encoder, the multi-modal interaction process runs through the entire network. And most of the previous global attention methods~\cite{wang2019asymmetric,hu2020bi} need to calculate an affinity matrix with the size of $HW \times HW$, so increasing the size of the input image will lead to a sharp increase in computational cost. The size of the affinity matrix calculated in VLMG is much smaller than $HW \times HW$ (e.g. $(HW) \times T$ in Eq.~\ref{v_to_l} and Eq.~\ref{l_to_v}, $T \times T$ in Eq.~\ref{transformer_block}). This also makes our encoder more suitable for real applications.
\subsection{Language-Guided Multi-Scale Dynamic Filtering}
For referring video segmentation, the language expression contains important spatial-temporal information related to the video. Therefore, we can use it to guide the cross-frame feature fusion.
Let $\overline{\mathrm{V}}^r \in \mathbb{R}^{C \times H \times W}$ and $\overline{\mathrm{V}}^c \in \mathbb{R}^{C \times H \times W}$ indicate the features of the reference frame and the current frame, respectively. They are generated by Eq.~\ref{multi_out}.
To capture useful spatial-temporal information from the reference frame and the current frame through the guidance of linguistic features,
we firstly calculate the affinity matrix ${\mathrm{A}_{\scriptscriptstyle r}} \in \mathbb{R}^{(HW)\times T}$ between $\overline{\mathrm{V}}^r$ and $\mathrm{L}_0$, as:
\begin{equation}
\begin{split}
& {\mathrm{A}_{\scriptscriptstyle r}}= {\rm softmax}(({\mathrm{W}_v^3 \overline{\mathrm{V}}^r)}^\top  ({ \mathrm{W}_l^3 \mathrm{L}_0) }).\\
\end{split}   \label{lag_ref}
\end{equation}
The column-wise normalized similarity matrix is used to map the reference feature $\overline{\mathrm{V}}^r$ to the language space, as:
\begin{equation}
\begin{split}
& \widetilde{\mathrm{V}}^r = \overline{\mathrm{V}}^r {\mathrm{A}_{\scriptscriptstyle r}} .\\
\end{split}
\end{equation}
We further adopt the transformer block to mix $\widetilde{\mathrm{V}}^r \in \mathbb{R}^{C \times T}$ and $\mathrm{L}_0 \in \mathbb{R}^{C \times T}$ and obtain the reference-frame modulated language feature $\mathrm{L}^{r} \in \mathbb{R}^{C \times T}$:
\begin{equation}
\begin{split}
& \widehat{\mathrm{L}}^r = \mathrm{MSA}(\mathrm{LN}(\widetilde{\mathrm{V}}^r + \mathrm{L}_0)) + (\widetilde{\mathrm{V}}^r + \mathrm{L}_0),\\
& \mathrm{L}^{r} = \mathrm{MLP}(\mathrm{LN}(\widehat{\mathrm{L}}^r)) + \widehat{\mathrm{L}}^r.
\end{split}
\end{equation}

Then, we superpose the current frame $\overline{\mathrm{V}}^c$ to further compute a multi-frame modulated language feature $\mathrm{L}^{c} \in \mathbb{R}^{C \times T}$ as follows:
\begin{equation}
\begin{split}
& \widetilde{\mathrm{V}}^c= \overline{\mathrm{V}}^c ({\rm softmax}(({\mathrm{W}_v^4 \overline{\mathrm{V}}^c)}^\top  ({ \mathrm{W}_l^4 \mathrm{L}^r)) }),\\
& \widehat{\mathrm{L}}^c = \mathrm{MSA}(\mathrm{LN}(\widetilde{\mathrm{V}}^c + \mathrm{L}^r)) + (\widetilde{\mathrm{V}}^c + \mathrm{L}^r),\\
& \mathrm{L}^{c} = \mathrm{MLP}(\mathrm{LN}(\widehat{\mathrm{L}}^c)) + \widehat{\mathrm{L}}^c.
\end{split}  \label{dynamic_lan_guided}
\end{equation}
Thus, under the guidance of language, the feature $\mathrm{L}^{c}$ successively fuses the language related spatial-temporal information from the reference frame and the current frame.

In order to generate the position-specific dynamic filters to accurately calibrate the current-frame feature based on the inter-frame and inter-modality information, we in advance learn the position-adaptive guidance feature based on the feature $\mathrm{L}^{c}$:
\begin{equation}
\begin{split}
& \overrightharpoonup{{\mathrm{V}}}^c= \mathrm{L}^c ({\rm softmax}((({\mathrm{W}_v^5 \overline{\mathrm{V}}^c)}^\top  ({ \mathrm{W}_l^5 \mathrm{L}^c))^\top) }),\\
\end{split}  \label{dynamic_adp}
\end{equation}
where $\overrightharpoonup{{\mathrm{V}}}^c \in \mathbb{R}^{C \times H \times W}$.
$\overrightharpoonup{{{v}}}^c_{i,j} \in \mathbb{R}^{C}$ and ${\overline{v}}^c_{i,j} \in \mathbb{R}^{C}$ represent the vectors of position $(i,j)$  in $\overrightharpoonup{{\mathrm{V}}}^c $ and $\overline{\mathrm{V}}^c$, respectively.
Then $\overrightharpoonup{{{v}}}^c_{i,j}$ is used to generate a set of dynamic kernels to filter the features of the neighborhood around $(i,j)$:
\begin{equation}
\begin{split}
& {v_d}' = \sum\limits_{k=-1}^{1} \sum\limits_{l=-1}^{1} (w_{k,l} \overrightharpoonup{{{v}}}^c_{i,j}) \odot {\overline{v}}^c_{i+k \times d,j+l \times d},\\
\end{split} \label{dynamic}
\end{equation}
where $w_{k,l} \in \mathbb{R}^{C \times C}$ is the learnable parameter. $\odot$ denotes the element-wise multiplication. Actually, Eq.\ref{dynamic} can be understood as a 3$\times$3 depth-wise convolution with dilated rate $d$. During the process of filtering, the multi-scale neighborhood is adopted. We then concatenate $\overline{v}^c$ and multiple ${v_d}'$ of different dilated rates and follow a convolution layer to combine them. Finally, the resulted cross-frame multi-modal features from stage3$\sim$stage5 and the visual features from stage1$\sim$stage2 are feed into the decoder (feature pyramid network) to predict the final segmentation result.

\noindent\textbf{Comparison with Other Temporal Models.} Our LMDF is different from other temporal modeling methods, e.g. CMDy~\cite{wang2020context} and LGFS~\cite{hui2021collaborative}.
1) They exploit the global guidance features of language after max or average pooling to generate the filters. The pooling operation causes all spatial positions of visual feature to share the same language guidance. While the LMDF combines the inter-modality and inter-frame information to yield the position-adaptive language guidance for dynamic filtering, which is more flexible and adaptive to appearance changes.
2) The LMDF learns the frame-specific filters, and implements the cross-modal and cross-frame interaction twice. The \emph{pre-interaction} process before dynamic kernel generation sequentially mixes the language features with the multi-modal features of reference frame and current frame, which can effectively obtain spatial-temporal guidance information. Next, the \emph{re-interaction} process of dynamic filtering further guarantees the temporal coherence of features for segmentation prediction. While LGFS~\cite{hui2021collaborative} uses the global language features to generate query-specific filters, which are shared along the whole video. And, it achieves the cross-modal and cross-frame interaction once. The CMDy~\cite{wang2020context} uses 3D convolution to obtain the multi-frame mixed feature for each frame. The mixed temporal information without language guidance may confuse spatial features of the current frame.
\begin{table*}[t]
\setlength{\tabcolsep}{4pt}
\small
\centering
\caption{\small{Quantitative evaluation on A2D Sentences. - denotes no data available. $^*$ denotes utilizing additional optical flow input.}} 
\renewcommand{\arraystretch}{1.0}
\begin{tabular}{p{3.0cm}<{\centering}||p{1.2cm}<{\centering}|p{1.2cm}<{\centering}|p{1.2cm}<{\centering}|p{1.2cm}<{\centering}
|p{1.2cm}<{\centering}|p{1.2cm}<{\centering}|p{1.2cm}<{\centering}|p{1.2cm}<{\centering}}
\hline
\multirow{2}{*}{Method}
&\multicolumn{5}{c|}{Precision} & \multicolumn{1}{c|}{mAP} & \multicolumn{2}{c}{IoU}\\
\cline{2-9}
&prec@0.5 &prec@0.6 &prec@0.7 &prec@0.8 &prec@0.9   &0.5:0.95   &Overall &Mean  \\
\hline \hline
Hu \emph{et al}.$_{16}$~\cite{hu2016segmentation}               &34.8 &23.6 &13.3 &3.3  &0.1  &13.2 &47.4 &35.0\\
Li \emph{et al}.$_{17}$~\cite{li2017tracking}                   &38.7 &29.0 &17.5 &6.6  &0.1  &16.3 &51.5 &35.4\\
Gavrilyuk \emph{et al}.$_{18}$~\cite{gavrilyuk2018actor}        &47.5 &34.7 &21.1 &8.0  &0.2  &19.8 &53.6 &42.1\\
Gavrilyuk \emph{et al}.$_{18}$~\cite{gavrilyuk2018actor}$^*$    &50.0 &37.6 &23.1 &9.4  &0.4  &21.5 &55.1 &42.6\\
ACGA$_{19}$~\cite{wang2019asymmetric}                           &55.7 &45.9 &31.9 &16.0 &2.0  &27.4 &60.1 &49.0\\
VT-Capsule$_{20}$~\cite{mcintosh2020visual}                     &52.6 &45.0 &34.5 &20.7 &3.6  &30.3 &56.8 &46.0\\
CMDy$_{20}$~\cite{wang2020context}                              &60.7 &52.5 &40.5 &23.5 &4.5  &33.3 &62.3 &53.1\\
PRPE$_{20}$~\cite{ningpolar}                                    &63.4 &57.9 &48.3 &32.2 &8.3  &38.8 &66.1 &52.9\\
CMSA-V$_{21}$~\cite{ye2021referring}                            &48.7 &43.1 &35.8 &23.1 &5.2  &-    &61.8 &43.2\\
LGFS$_{21}$~\cite{hui2021collaborative}                         &65.4 &58.9 &49.7 &33.3 &9.1  &39.9 &66.2 &56.1\\
CMPC-V$_{21}$~\cite{liu2021cross}                               &65.5 &59.2 &50.6 &34.2 &9.8  &40.4 &65.3 &57.3\\
\hline \hline
Ours                                                            &70.2 &66.3 &58.5 &42.8 &15.1 &46.9 &71.4 &59.8\\ 
\hline
\end{tabular}
\label{tab:video_a2d}
\end{table*}
\begin{table*}[t]
\setlength{\tabcolsep}{4pt}
\small
\centering
\caption{\small{Quantitative evaluation on J-HMDB Sentences. - denotes no data available.}} 
\renewcommand{\arraystretch}{1.0}
\begin{tabular}{p{3.0cm}<{\centering}||p{1.2cm}<{\centering}|p{1.2cm}<{\centering}|p{1.2cm}<{\centering}|p{1.2cm}<{\centering}
|p{1.2cm}<{\centering}|p{1.2cm}<{\centering}|p{1.2cm}<{\centering}|p{1.2cm}<{\centering}}
\hline
\multirow{2}{*}{Method}
&\multicolumn{5}{c|}{Precision} & \multicolumn{1}{c|}{mAP} & \multicolumn{2}{c}{IoU}\\
\cline{2-9}
&prec@0.5 &prec@0.6 &prec@0.7 &prec@0.8 &prec@0.9   &0.5:0.95   &Overall &Mean  \\
\hline \hline
Hu \emph{et al}.$_{16}$~\cite{hu2016segmentation}               &63.3 &35.0 &8.5  &0.2  &0.0 &17.8 &54.6 &52.8\\
Li \emph{et al}.$_{17}$~\cite{li2017tracking}                   &57.8 &33.5 &10.3 &0.6  &0.0 &17.3 &52.9 &49.1\\
Gavrilyuk \emph{et al}.$_{18}$~\cite{gavrilyuk2018actor}        &69.9 &46.0 &17.3 &1.4  &0.0 &23.3 &54.1 &54.2\\
ACGA$_{19}$~\cite{wang2019asymmetric}                           &75.6 &56.4 &28.7 &3.4  &0.0 &28.9 &57.6 &58.4\\
VT-Capsule$_{20}$~\cite{mcintosh2020visual}                     &67.7 &51.3 &28.3 &5.1  &0.0 &26.1 &53.5 &55.0\\
CMDy$_{20}$~\cite{wang2020context}                              &74.2 &58.7 &31.6 &4.7  &0.0 &30.1 &55.4 &57.6\\
PRPE$_{20}$~\cite{ningpolar}                                    &69.1 &57.2 &31.9 &6.0  &0.1 &29.4 &-    &-\\
CMSA-V$_{21}$~\cite{ye2021referring}                            &76.4 &62.5 &38.9 &9.0  &0.1 &-    &62.8 &58.1\\
LGFS$_{21}$~\cite{hui2021collaborative}                         &78.3 &63.9 &37.8 &7.6  &0.0 &33.5 &59.8 &60.4\\
CMPC-V$_{21}$~\cite{liu2021cross}                               &81.3 &65.7 &37.1 &7.0  &0.0 &34.2 &61.6 &61.7\\
\hline \hline
Ours                                                            &87.4 &79.1 &58.6 &18.2 &0.3 &44.1 &68.0 &66.6\\
\hline
\end{tabular}
\label{tab:video_HMDB}
\end{table*}
\section{Experimental Comparison}
\subsection{Datasets}
In order to illustrate the effectiveness of the proposed method, we conduct extensive experiments on four referring video segmentation datasets, which are A2D Sentences~\cite{gavrilyuk2018actor}, J-HMDB Sentences~\cite{gavrilyuk2018actor}, Refer-DAVIS2017~\cite{khoreva2018video}, and Refer-Youtube-VOS~\cite{seo2020urvos}.
Gavrilyuk \emph{et al.}~\cite{gavrilyuk2018actor} extended the original Actor-Action Dataset (A2D)~\cite{xu2015can} and J-HMDB~\cite{jhuang2013towards} datasets to the referring video segmentation task by adding additional natural language descriptions.
A2D Sentences is composed of 3,782 videos with 6,655 referring expressions. And this videos contain 8 different actions that are preformed by 7 actor classes. Following the previous setting~\cite{wang2019asymmetric,hui2021collaborative}, we split A2D into two parts, 3,017 of which are used for training and 737 for testing. Each video contains 3 to 5 frame pixel-level segmentation masks. For J-HMDB Sentences, it consists of 928 videos with 21 different action classes. Each video contains a corresponding language expression. And the main purpose of this dataset is to evaluate the generalization ability of the model. Generally, previous methods directly use the model trained on the A2D dataset to test all the videos in this dataset.
Similarly, Refer-DAVIS2017 is based on DAVIS2017~\cite{perazzi2016benchmark}. It contains 90 videos, 60 for training and 30 for testing.
Refer-Youtube-VOS is by far the largest dataset, it contains a total of 3,978 videos and about 15,000 language expressions.
\subsection{Implementation Details}
\noindent\textbf{Experiment setting.}
We exploit the pytorch platform to implement our network, and adopt a Nvidia RTX 3090 GPU to train and test it. During training, we use the SGD optimizer to optimize the whole network. And we set the initial learning rate, momentum and weight decay to 1e-3, 0.9 and 5e-4, respectively. The batch size is set to 16 and the input frame is resized to 320$\times$320. The maximum length of the referring expression is limited to 20. The dilated rates of LMDS are set $d=1,3,5$.
For Refer-DAVIS2017 and Refer-Youtube-VOS, we use the referring image segmentation dataset RefCOCO~\cite{yu2016modeling} to assist their training. And random affine transformation is used to for data augmentation. The maximum number of iterations for A2D Sentences, Refer-DAVIS2017, and Refer-Youtube-VOS are set to 75,000, 150,000, and 300,000, respectively. And after 50,000, 80,000, and 200,000 iterations, the learning rate is reduced by 5 times. J-HMDB Sentences is only used to verify the generalization ability of the model.

\noindent\textbf{Evaluation metrics.}
Following previous works~\cite{seo2020urvos,hui2021collaborative}, we take Prec@X, Overall IoU, mAP, Mean IoU (mean region similarity $\mathcal(J)$), mean contour accuracy ($\mathcal{F}$), and the average of and $\mathcal{J}$ and $\mathcal{F}$ ($\mathcal{J}\&\mathcal{F}$) to evaluate our method. Where $\rm X \in \{0.5, 0.6, 0.7, 0.8, 0.9\}$.
\subsection{Comparison with State-of-the-arts}
\begin{figure*}[htp]
\begin{center}
\begin{tabular}{c@{\hspace{0.6mm}}c@{\hspace{0.6mm}}c@{\hspace{1.8mm}}c@{\hspace{0.6mm}}c@{\hspace{0.6mm}}c@{\hspace{0.6mm}}c}
\multicolumn{3}{c}{ \rule{0pt}{10pt} {Query:  \emph{``a black colored''} }} &
\multicolumn{3}{c}{ \rule{0pt}{10pt} {Query:  \emph{``a big man on the right in a black jacket''}  }} \vspace{1mm}\\
\includegraphics[width=0.16\linewidth,height=0.10\linewidth]{fig/select_seg/1/480p_pigs_5_00000(0.9269)/img.png}&
\includegraphics[width=0.16\linewidth,height=0.10\linewidth]{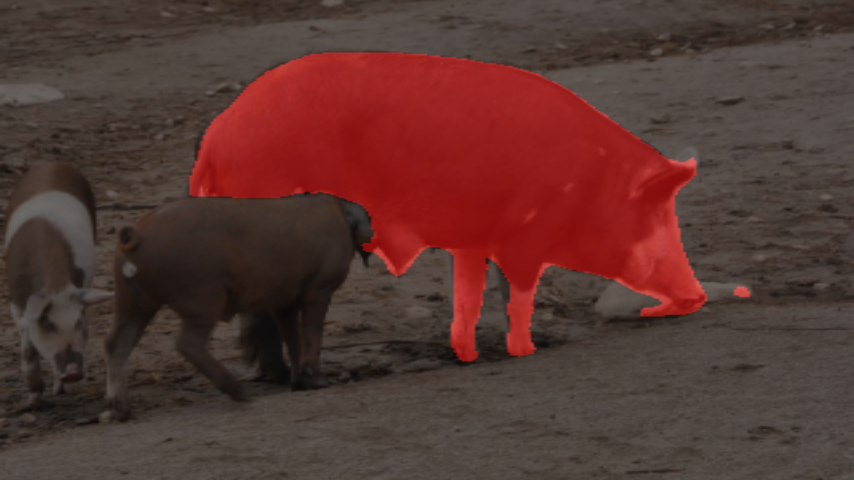}&
\includegraphics[width=0.16\linewidth,height=0.10\linewidth]{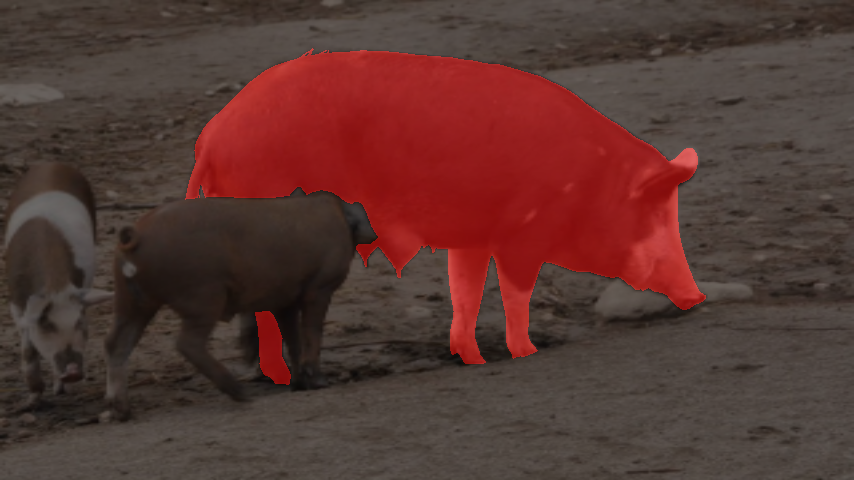}&
\includegraphics[width=0.16\linewidth,height=0.10\linewidth]{fig/select_seg/3/480p_loading_0_00000(0.9509)/img.png}&
\includegraphics[width=0.16\linewidth,height=0.10\linewidth]{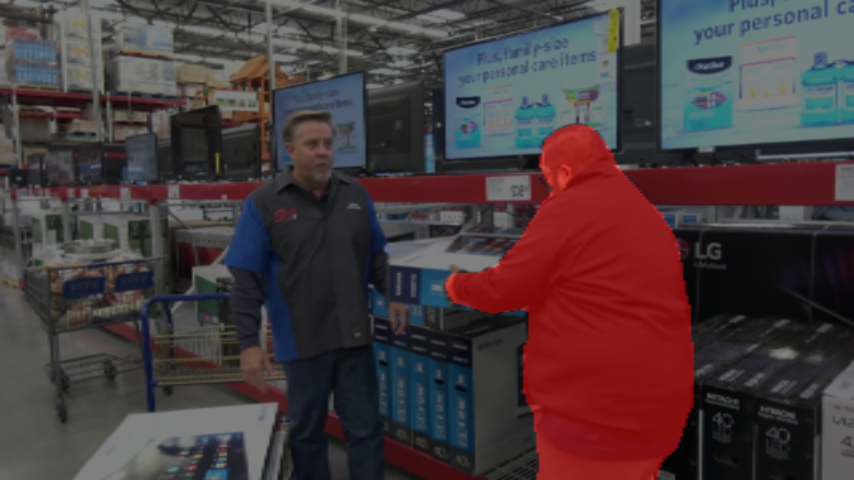}&
\includegraphics[width=0.16\linewidth,height=0.10\linewidth]{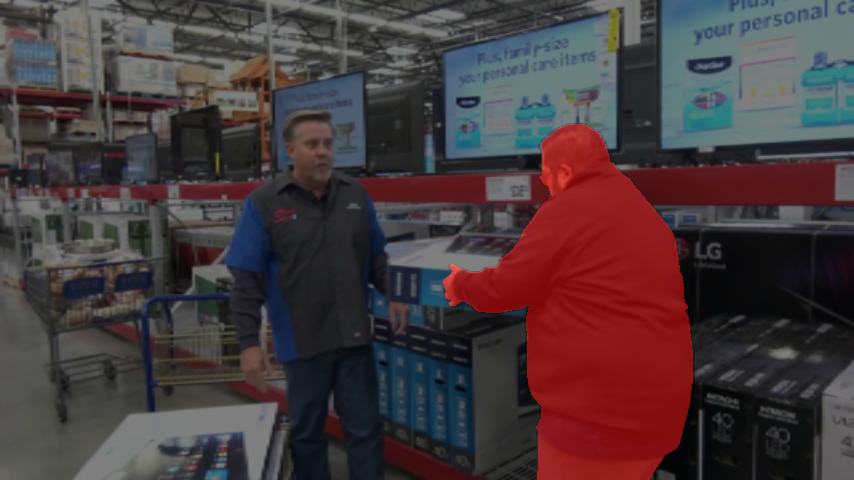}\vspace{-0mm}\\
\includegraphics[width=0.16\linewidth,height=0.10\linewidth]{fig/select_seg/1/480p_pigs_5_00035(0.9251)/img.png}&
\includegraphics[width=0.16\linewidth,height=0.10\linewidth]{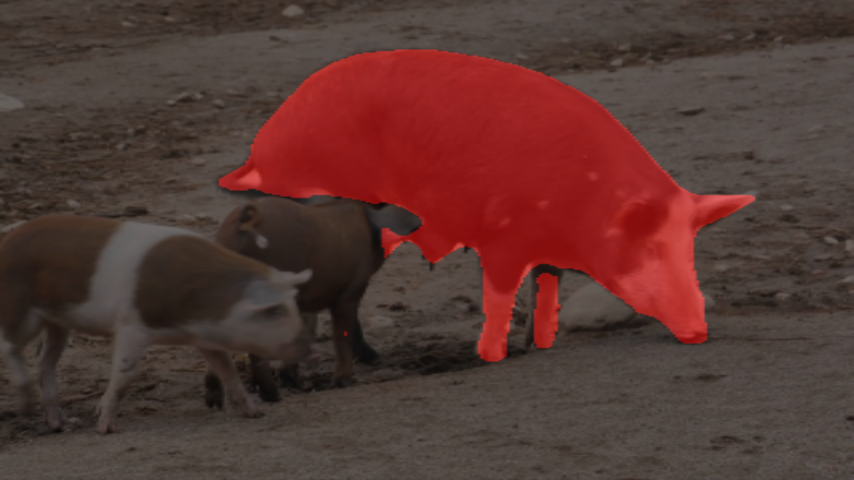}&
\includegraphics[width=0.16\linewidth,height=0.10\linewidth]{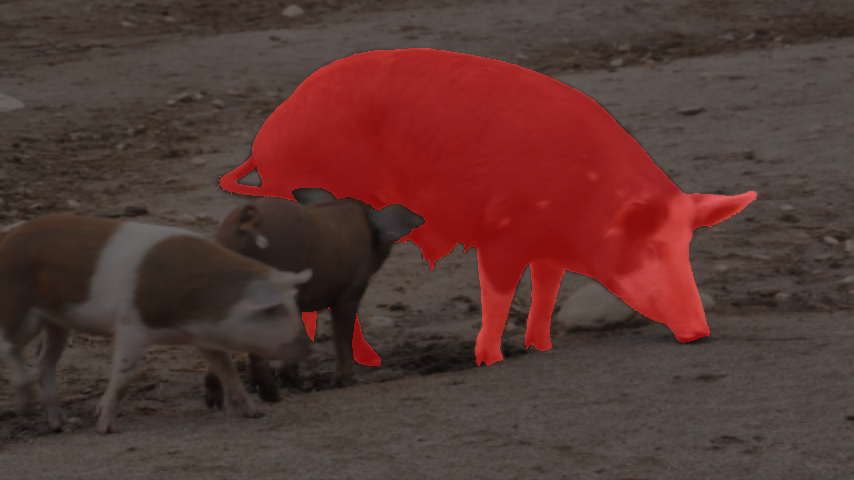}&
\includegraphics[width=0.16\linewidth,height=0.10\linewidth]{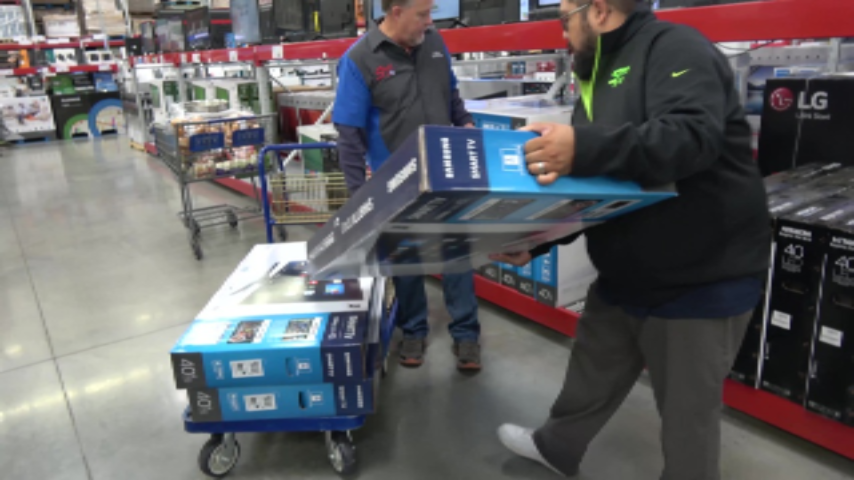}&
\includegraphics[width=0.16\linewidth,height=0.10\linewidth]{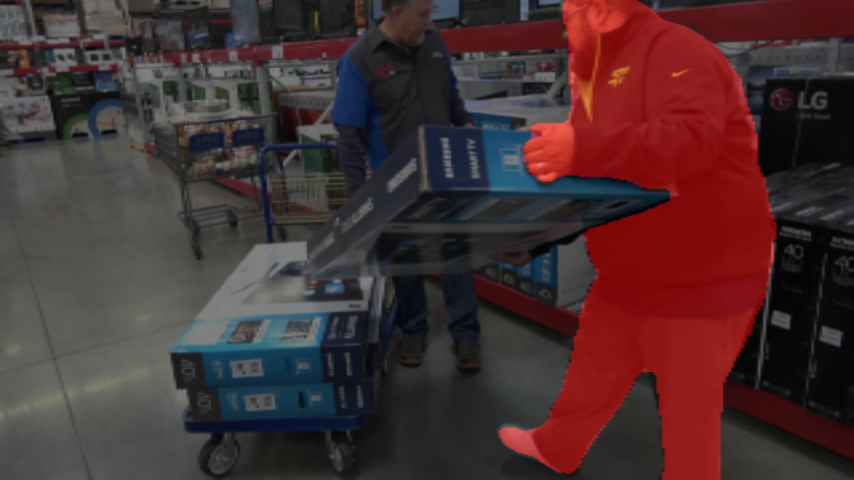}&
\includegraphics[width=0.16\linewidth,height=0.10\linewidth]{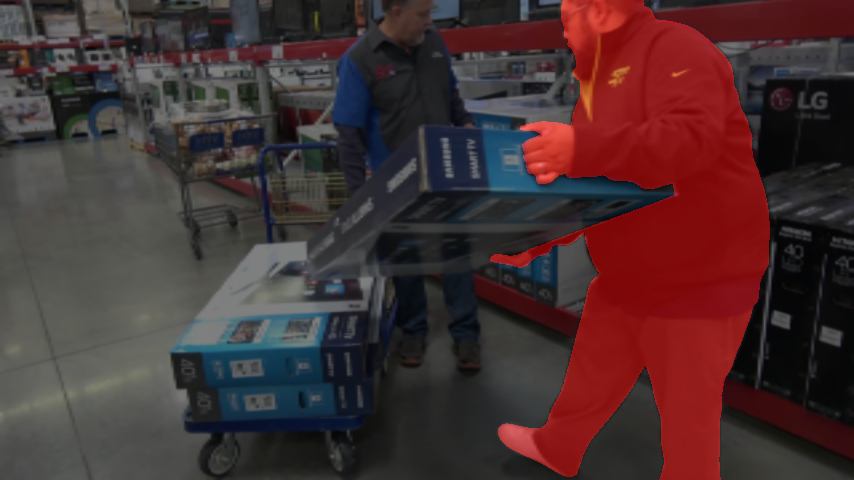}\vspace{-0mm}\\
\includegraphics[width=0.16\linewidth,height=0.10\linewidth]{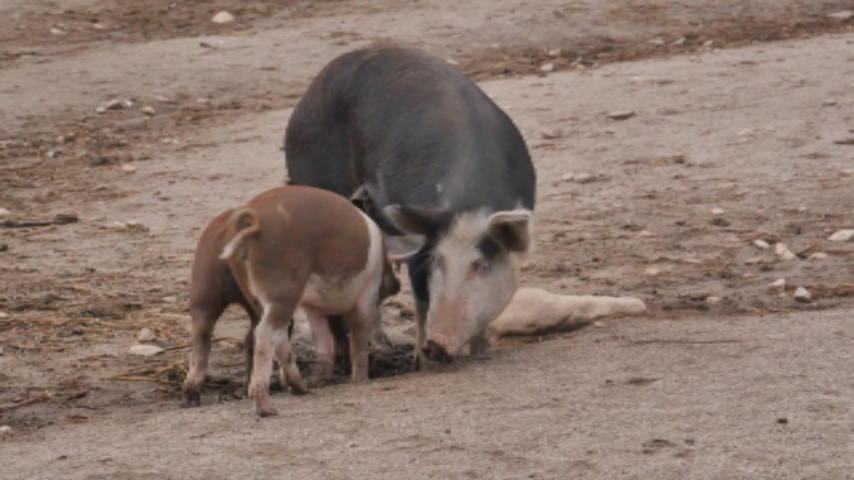}&
\includegraphics[width=0.16\linewidth,height=0.10\linewidth]{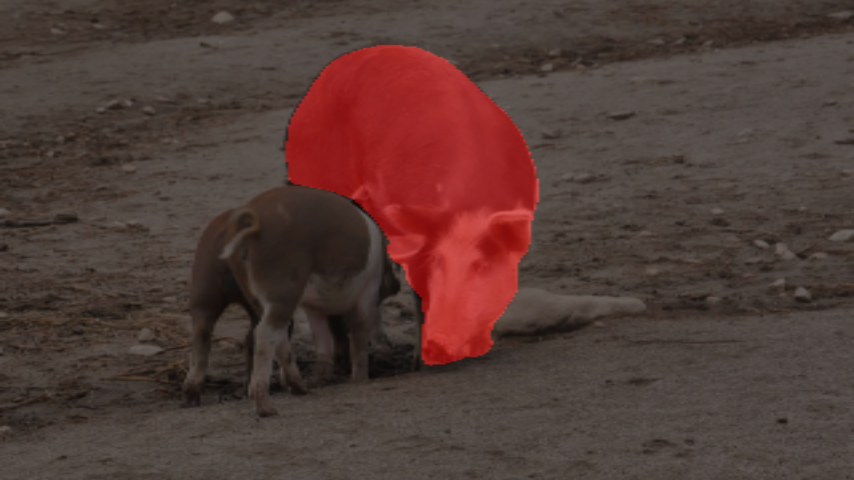}&
\includegraphics[width=0.16\linewidth,height=0.10\linewidth]{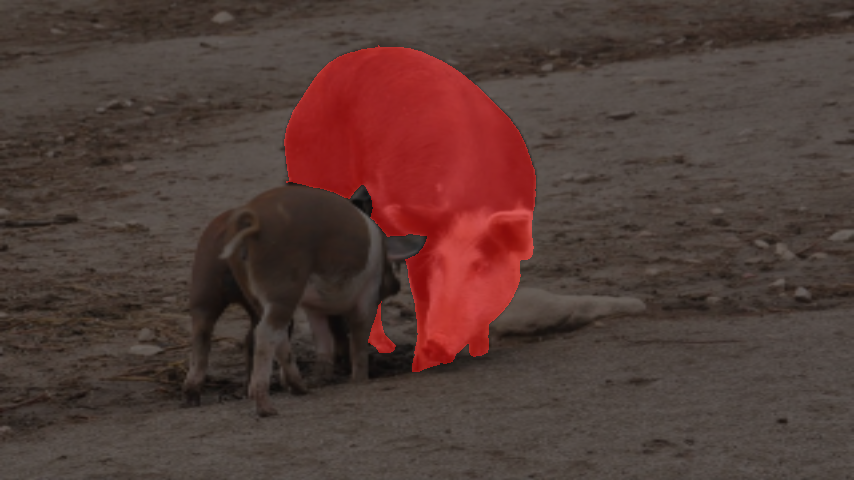}&
\includegraphics[width=0.16\linewidth,height=0.10\linewidth]{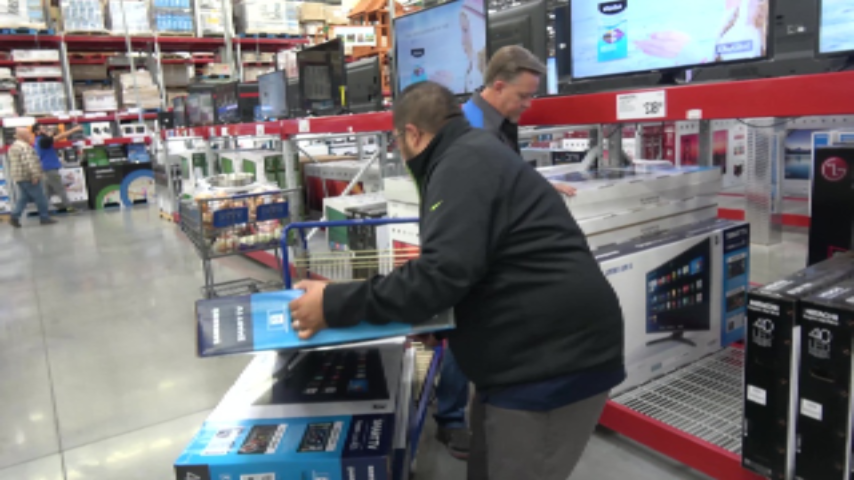}&
\includegraphics[width=0.16\linewidth,height=0.10\linewidth]{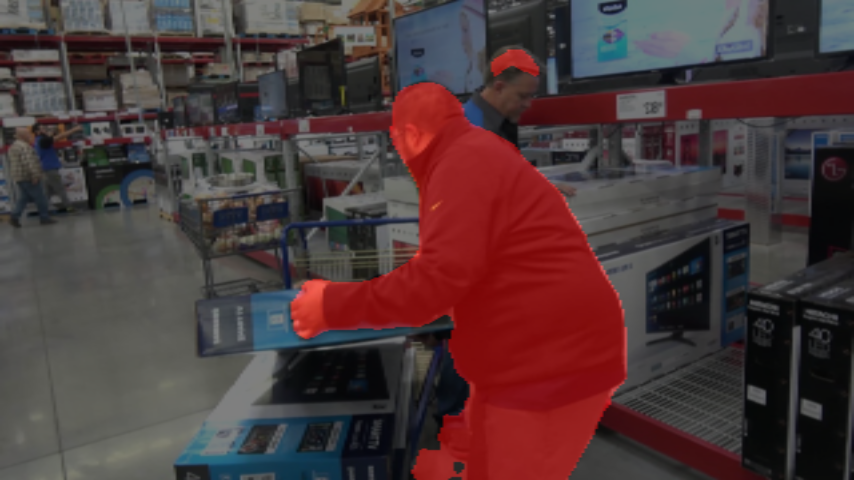}&
\includegraphics[width=0.16\linewidth,height=0.10\linewidth]{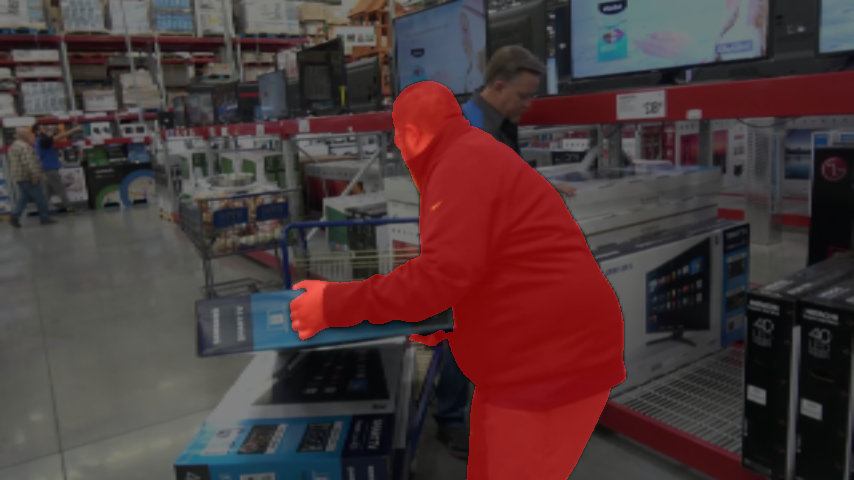}\vspace{-0mm}\\
\fontsize{8.0pt}{\baselineskip}\selectfont Image&
\fontsize{8.0pt}{\baselineskip}\selectfont Result&
\fontsize{8.0pt}{\baselineskip}\selectfont GT&
\fontsize{8.0pt}{\baselineskip}\selectfont Image&
\fontsize{8.0pt}{\baselineskip}\selectfont Result&
\fontsize{8.0pt}{\baselineskip}\selectfont GT\\
\end{tabular}
\end{center}
\vspace{-5mm}
\caption{\small{Visual examples of referring video segmentation.}}\label{fig:performance}
\vspace{0mm}
\end{figure*}

\begin{table}[t]
\setlength{\tabcolsep}{4pt}
\small
\centering
\caption{Quantitative evaluation on Refer-DAVIS2017 val.} 
\renewcommand{\arraystretch}{1.0}
\begin{tabular}{p{3cm}<{\centering}||p{1.0cm}<{\centering}|p{1.0cm}<{\centering}|p{1.0cm}<{\centering}}
\hline
\multirow{1}{*}{Mtehod}          &$\mathcal{J}$    &$\mathcal{F}$  &$\mathcal{J} \& \mathcal{F}$\\
\hline \hline
Khoreva \emph{et al}.$_{18}$~\cite{khoreva2018video}        &37.3   &41.3  &39.3\\
URVOS$_{20}$~\cite{seo2020urvos}                            &41.23  &47.01 &44.12\\
Ours                                                        &47.71  &52.33 &50.02\\
\hline
\end{tabular}
\label{tab:video_davis}
\end{table}
We compare the performance of our method and the previous state-of-the-art methods on four different datasets.
First of all, for the A2D Sentences dataset, we can find that the proposed method is significantly better than the previous methods. This also reflects the effectiveness of the two-stream encoder and language-guided dynamic filtering.
For all the evaluation metrics in Tab.~\ref{tab:video_a2d}, our model achieves an average absolute gain of more than 5.5\%. Especially on Prec@0.6, Prec@0.7 and Prec@0.8, our method obtains absolute enhancement of 7.1\%, 7.9\% and 8.6\% compared with the second-best method. For the metric Prec@0.9, our method improves it from 9.8\% to 15.1\%, which is 1.5 times the previous performance. This also proves that our network can more accurately perceive the boundary of the object.
%
\begin{table}[t]
\setlength{\tabcolsep}{4pt}
\small
\centering
\caption{Quantitative evaluation on Refer-Youtube-VOS val.} 
\renewcommand{\arraystretch}{1.0}
\begin{tabular}{p{3cm}<{\centering}||p{1.0cm}<{\centering}|p{1.0cm}<{\centering}|p{1.0cm}<{\centering}}
\hline
\multirow{1}{*}{Mtehod}          &$\mathcal{J}$    &$\mathcal{F}$  &$\mathcal{J} \& \mathcal{F}$\\
\hline \hline
URVOS$_{20}$~\cite{seo2020urvos}             &45.27         &49.19        &47.23\\
CMPC-V$_{21}$~\cite{liu2021cross}            &45.64         &49.32        &47.48\\
Ours                                         &48.44         &50.67        &49.56\\
\hline
\end{tabular}
\label{tab:video_youtube}
\end{table}
Following the previous methods~\cite{hui2021collaborative,liu2021cross}, we further use the J-HMDB Sentences dataset to verify the generalization ability of the model trained on the A2D Sentences dataset. As shown in Tab.~\ref{tab:video_HMDB}, we can find that the proposed model achieves the best performance under all evaluation metrics. Particularly, our network shows a very significant improvement of 19.7\% on Prec@0.7.
For the Refer-DAVIS2017, it only contains 60 training videos, so the previous methods~\cite{khoreva2018video,seo2020urvos} usually take advantage of RefCOCO~\cite{yu2016modeling} for data augmentation. Our model achieves the gain of 6.5\% and 5.3\% in terms of $\mathcal{J}$ and $\mathcal{F}$.
Finally, Tab.~\ref{tab:video_youtube} illustrates the accuracy on the large-scale dataset Refer-Youtube-VOS. Our model surpasses other methods in all evaluation metrics.
We also present some visual cases in Fig.~\ref{fig:performance}. It can be seen that our method can produce accurate segmentation masks, even when the language expression does not contain location information or the length of query is various.
\begin{table*}[t]
\setlength{\tabcolsep}{4pt}
\small
\centering
\caption{\small{Ablation study on the Refer-DAVIS2017 val.}} 
\renewcommand{\arraystretch}{1.0}
\begin{tabular}{c|cccc||p{1.2cm}<{\centering}|p{1.2cm}<{\centering}|p{1.2cm}<{\centering}|p{1.2cm}<{\centering}|p{1.2cm}<{\centering}|p{1.2cm}<{\centering}|p{1.2cm}<{\centering}|p{1.2cm}<{\centering}}
\hline
\multirow{1}{*}{} &\multirow{1}{*}{EFN} &\multirow{1}{*}{EFN$^*$} &\multirow{1}{*}{Dual} &\multirow{1}{*}{LMDF}
&prec@0.5 &prec@0.6 &prec@0.7 &prec@0.8 &prec@0.9 &$\mathcal{J}$ &$\mathcal{F}$ &$\mathcal{J} \& \mathcal{F}$  \\

\hline \hline
\multirow{4}{*}{}
&\checkmark &           &           &           &43.83 &38.45 &32.05 &23.46 &10.20 &40.33 &45.67 &43.00           \\
&           &\checkmark &           &           &44.42 &38.62 &31.63 &23.67 &9.45  &41.68 &47.22 &44.45          \\
&           &           &\checkmark &           &50.38 &44.65 &37.11 &26.83 &10.40 &45.20 &50.19 &47.70          \\
&           &           &\checkmark &\checkmark &53.93 &47.57 &40.44 &30.06 &10.81 &47.71 &52.33 &50.02          \\
\hline \hline
\end{tabular}
\label{tab:ablation_video}
\end{table*}
\begin{table*}[t]
\setlength{\tabcolsep}{4pt}
\small
\centering
\caption{\small{Comparison of different settings of LMDF on the Refer-DAVIS2017 val dataset. $d$ denotes the dilated rate.}} 
\renewcommand{\arraystretch}{1.0}
\begin{tabular}{p{2cm}<{\centering}|p{2.5cm}<{\centering}|p{2.5cm}<{\centering}|p{2.5cm}<{\centering}|p{2.6cm}<{\centering}|p{2.6cm}<{\centering}}
\hline
\multirow{1}{*}{Metrics} &\multirow{1}{*}{$d=1,3,5$} &\multirow{1}{*}{$d=1$} &\multirow{1}{*}{MaxPool} &\multirow{1}{*}{w/o pre-interaction}
  &\multirow{1}{*}{w/ weight-sharing}   \\

\hline \hline
$\mathcal{J}$                 &47.71   &46.62   &46.41   &45.73   &46.50   \\
$\mathcal{F}$                 &52.33   &51.26   &51.68   &50.23   &51.05   \\
$\mathcal{J} \& \mathcal{F}$  &50.02   &48.94   &49.05   &47.98   &48.78   \\
\hline \hline
\end{tabular}
\label{tab:ablation_lmdf}
\end{table*}
\begin{figure*}[t]
\begin{center}
\begin{tabular}{c@{\hspace{0.5mm}}c@{\hspace{0.5mm}}c@{\hspace{0.5mm}}c@{\hspace{0.5mm}}c@{\hspace{0.5mm}}c@{\hspace{0.5mm}}c}
\multicolumn{6}{c}{ \rule{0pt}{10pt} {Query:  \emph{``a black swan"} }} \vspace{1mm}\\
\includegraphics[width=0.16\linewidth,height=0.10\linewidth]{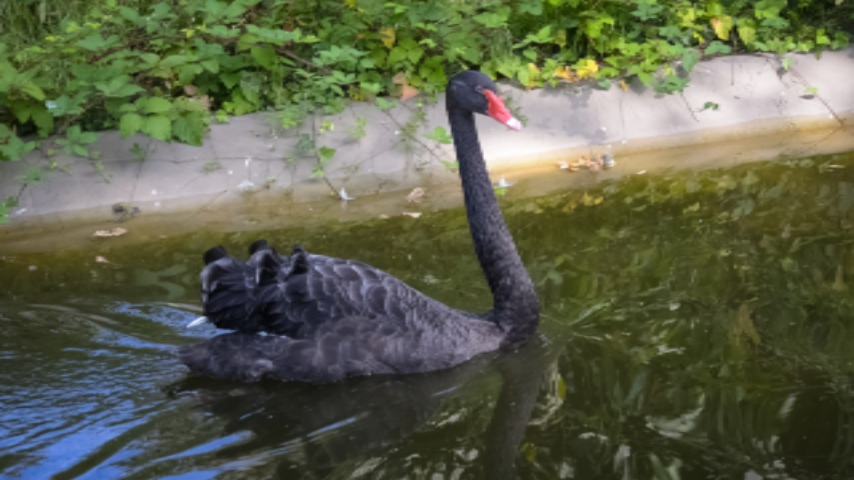}&
\includegraphics[width=0.16\linewidth,height=0.10\linewidth]{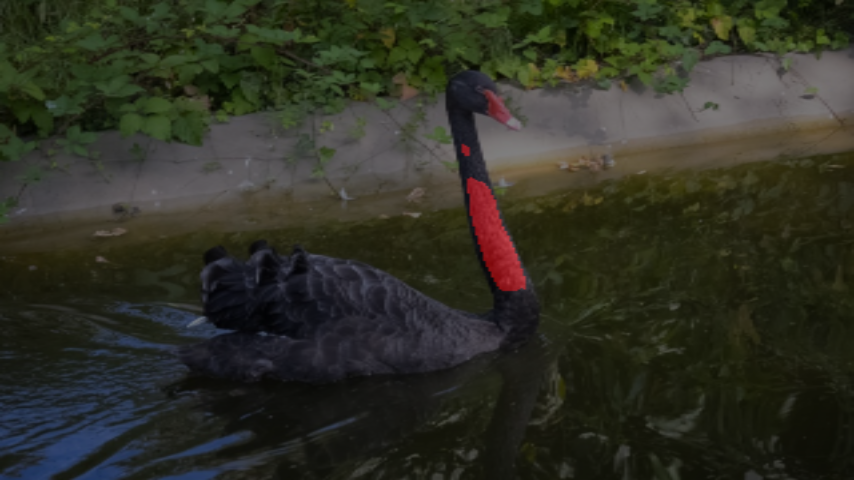}&
\includegraphics[width=0.16\linewidth,height=0.10\linewidth]{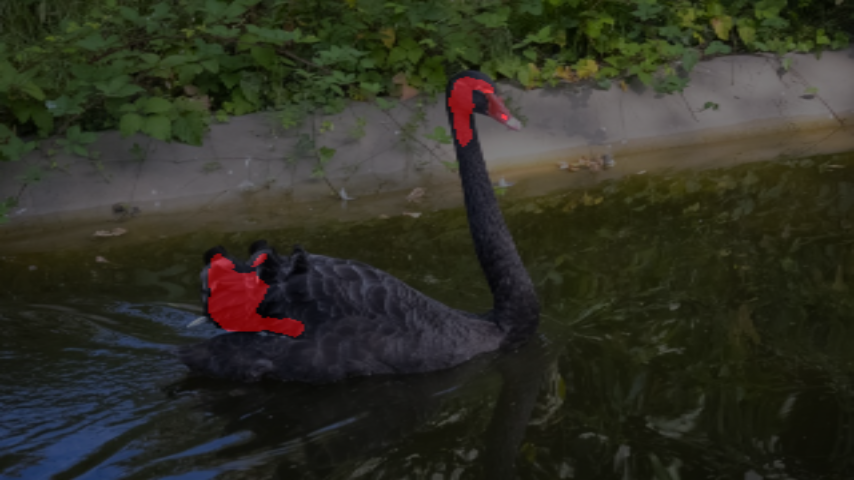}&
\includegraphics[width=0.16\linewidth,height=0.10\linewidth]{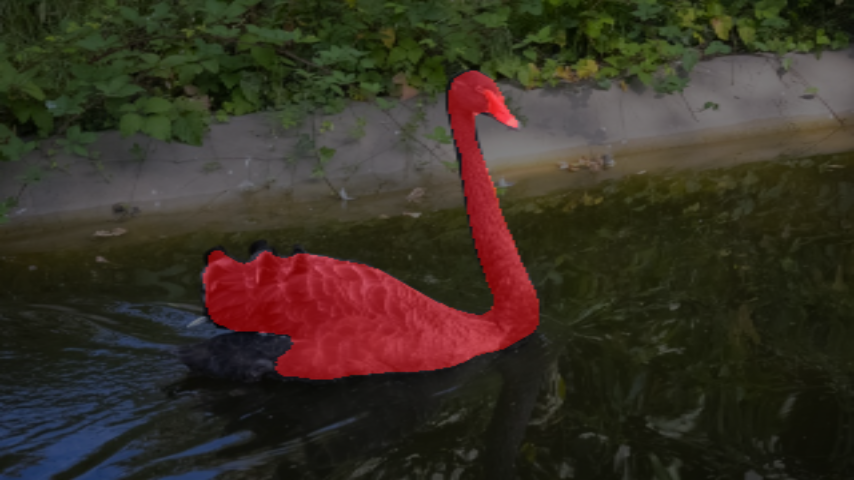}&
\includegraphics[width=0.16\linewidth,height=0.10\linewidth]{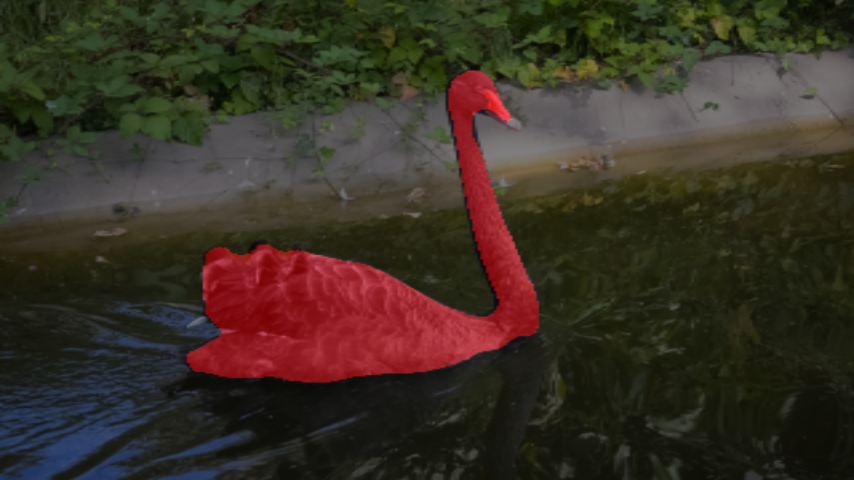}&
\includegraphics[width=0.16\linewidth,height=0.10\linewidth]{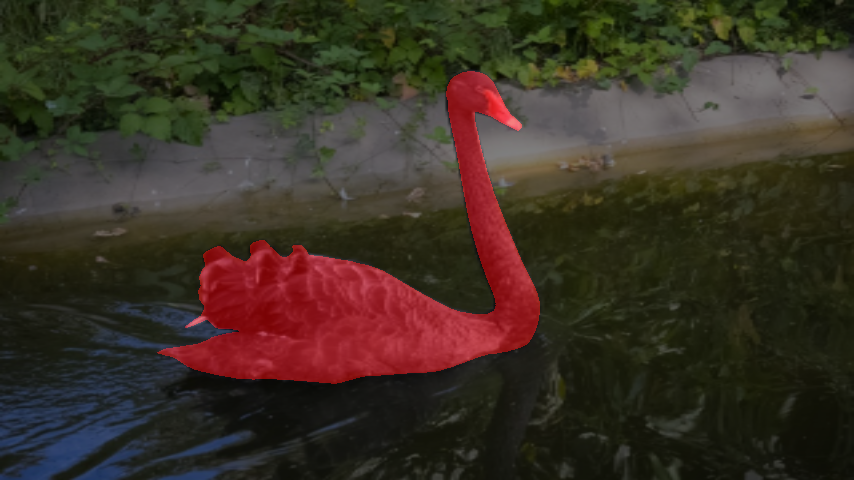}&\vspace{0mm}\\
\multicolumn{6}{c}{ \rule{0pt}{10pt} {Query:  \emph{``a girl on the left holding two cell phones"} }} \vspace{1mm}\\
\includegraphics[width=0.16\linewidth,height=0.10\linewidth]{fig/select_ablation/2/480p_lab-coat_4_00027_img_a_girl_on_the_left_holding_two_cell_phones.png}&
\includegraphics[width=0.16\linewidth,height=0.10\linewidth]{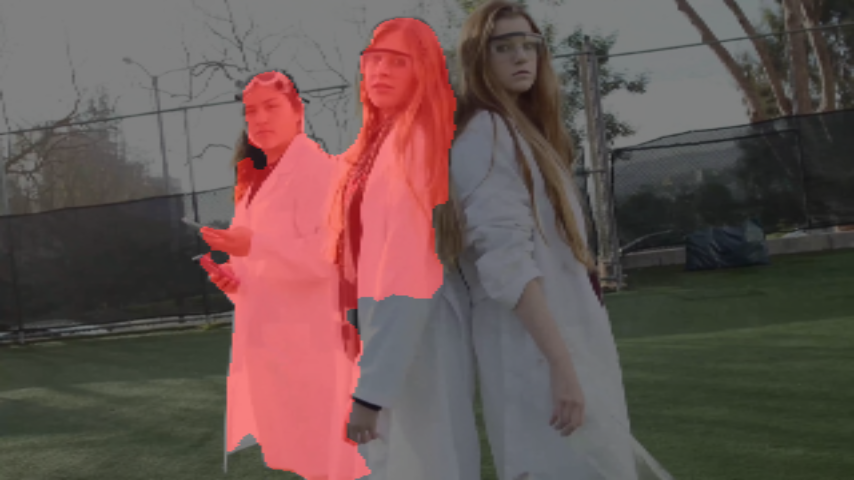}&
\includegraphics[width=0.16\linewidth,height=0.10\linewidth]{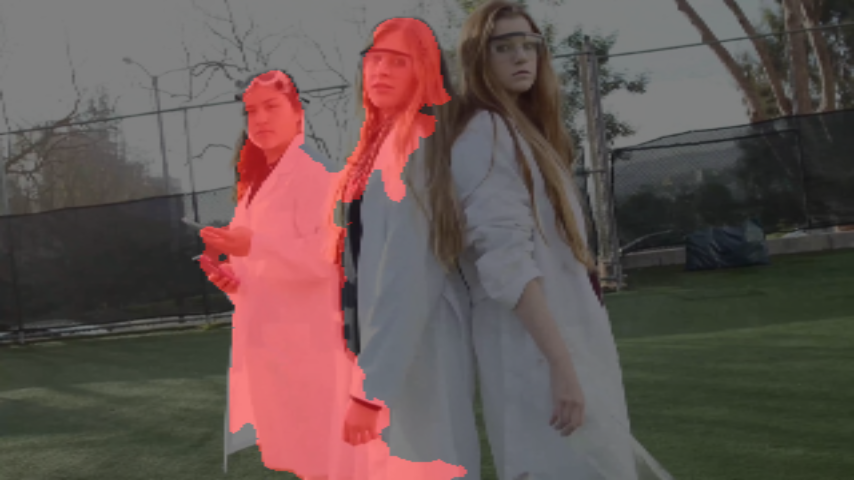}&
\includegraphics[width=0.16\linewidth,height=0.10\linewidth]{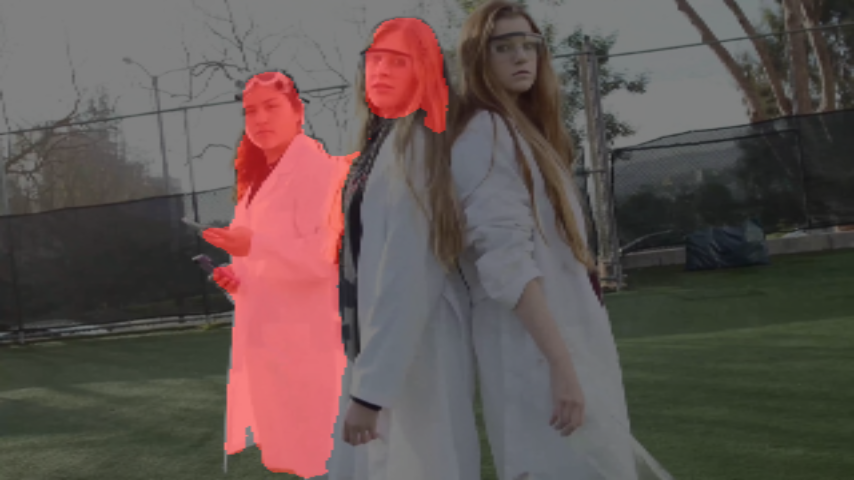}&
\includegraphics[width=0.16\linewidth,height=0.10\linewidth]{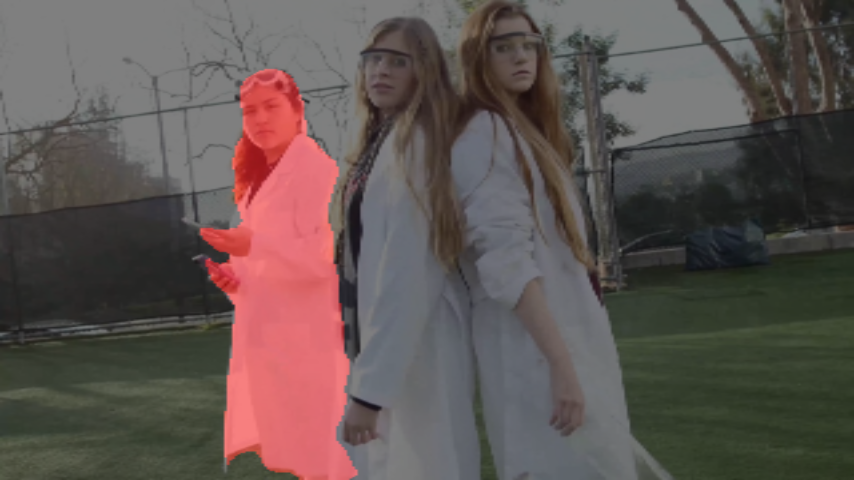}&
\includegraphics[width=0.16\linewidth,height=0.10\linewidth]{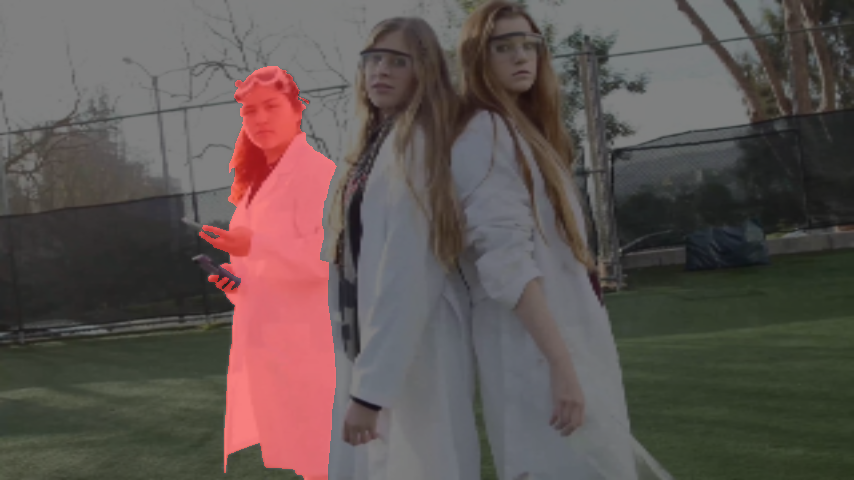}&\vspace{0mm}\\
\multicolumn{6}{c}{ \rule{0pt}{10pt} {Query:  \emph{``a blonde haired girl dancing in a blue dress"} }} \vspace{1mm}\\
\includegraphics[width=0.16\linewidth,height=0.10\linewidth]{fig/select_ablation/3/480p_dance-twirl_1_00077_img_a_blonde_haired_girl_dancing_in_a_blue_dress.png}&
\includegraphics[width=0.16\linewidth,height=0.10\linewidth]{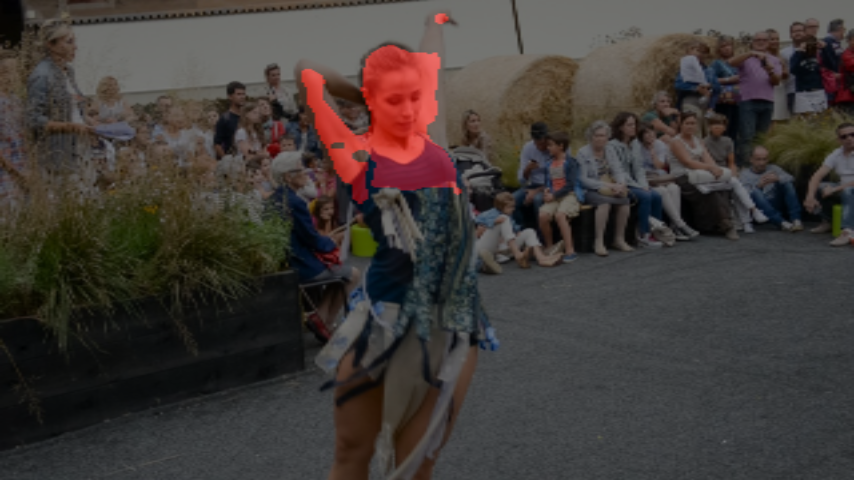}&
\includegraphics[width=0.16\linewidth,height=0.10\linewidth]{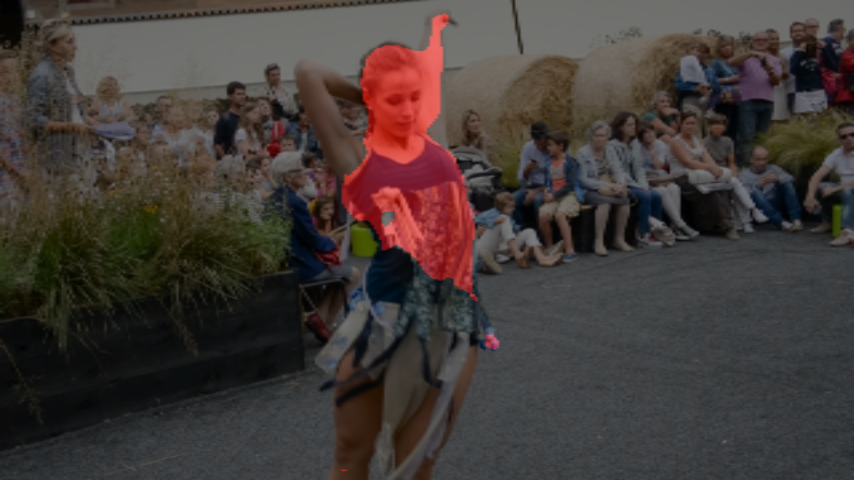}&
\includegraphics[width=0.16\linewidth,height=0.10\linewidth]{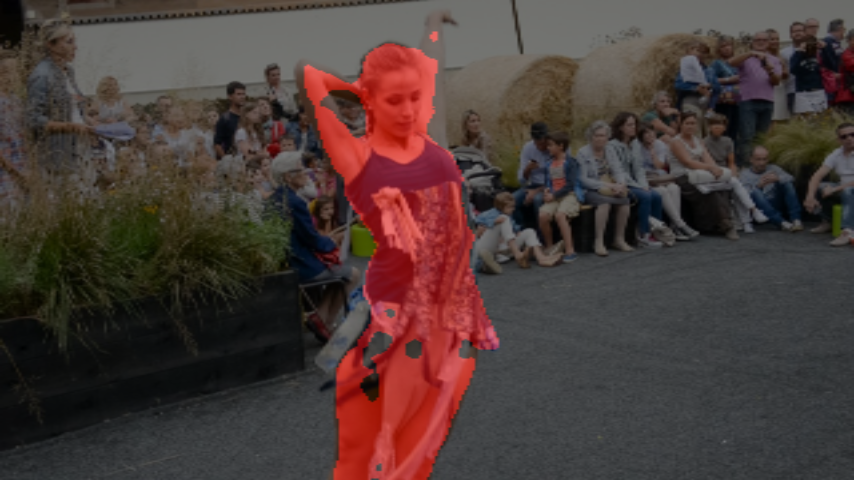}&
\includegraphics[width=0.16\linewidth,height=0.10\linewidth]{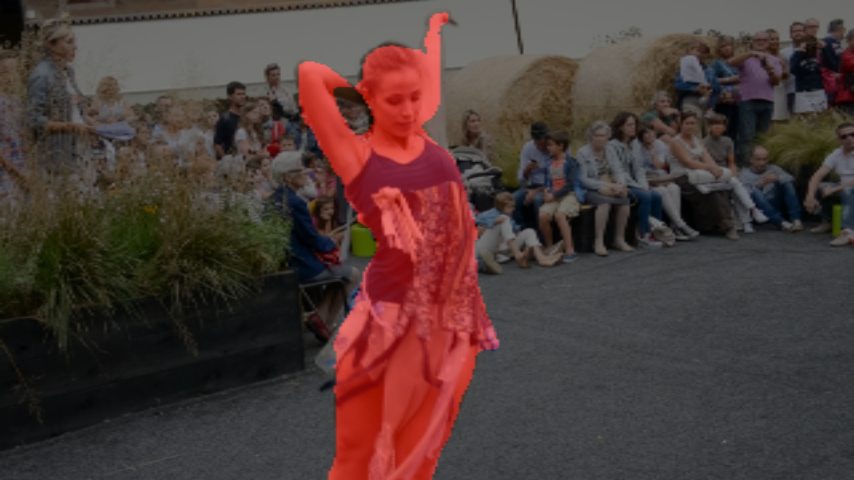}&
\includegraphics[width=0.16\linewidth,height=0.10\linewidth]{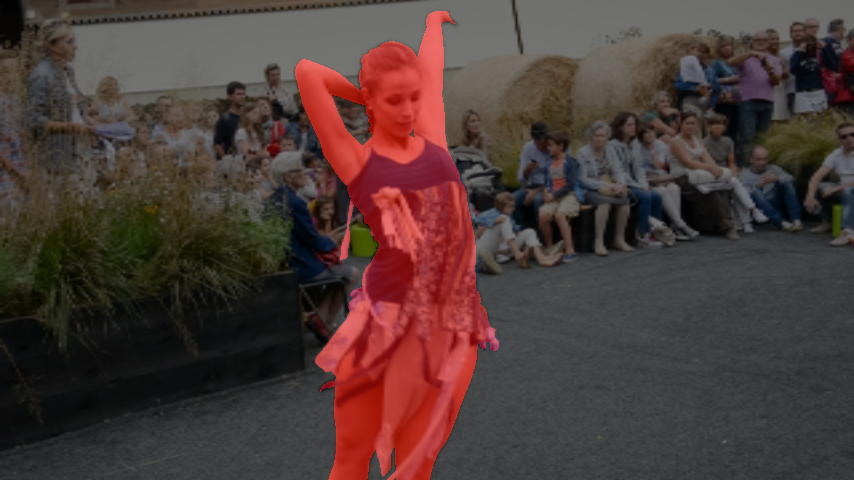}&\vspace{0mm}\\
\fontsize{8.0pt}{\baselineskip}\selectfont Image&
\fontsize{8.0pt}{\baselineskip}\selectfont EFN&
\fontsize{8.0pt}{\baselineskip}\selectfont EFN$^*$&
\fontsize{8.0pt}{\baselineskip}\selectfont Dual&
\fontsize{8.0pt}{\baselineskip}\selectfont Dual+LMDF&
\fontsize{8.0pt}{\baselineskip}\selectfont GT\\
\end{tabular}
\end{center}
\vspace{-5mm}
\caption{Visual examples of the proposed modules.}\label{fig:ablation_1}
\vspace{0mm}
\end{figure*}
\subsection{Ablation Study}
We carefully evaluate the proposed components on the Refer-DAVIS2017 dataset. The quantitative results are shown in Tab.~\ref{tab:ablation_video} and Tab.~\ref{tab:ablation_lmdf}.

\noindent\textbf{Vision-Language Interleaved Encoder.}
We first remove the vision-language mutual guidance module (VLMG) and language-guided multi-scale dynamic filtering module (LMDF) from the overall network to construct the basic encoder fusion network (EFN). In EFN, the initial vector representation ($\mathrm{L}_0$) of the language is directly inserted into the visual encoder three times by tile and convolution to achieve the fusion of language and vision (as used in~\cite{feng2021encoder}). In Tab~\ref{tab:ablation_video}, EFN$^*$ indicates that the transformer structure is used to extract the multi-granularity language context of $\mathrm{L}_0$. By comparing EFN$^*$ and EFN, we can find that the hierarchical language information in EFN$^*$ can be better integrated with the multi-level visual information, thereby improving the performance of the network. Next, we introduce VLMG to realize the mutual interweaving of vision and language (marked as Dual in Tab.~\ref{tab:ablation_video}). The experimental results show that the dual interleaved encoding strategy can further bring 3.5\%, 3.0\%, and 3.3\% gains in terms of $\mathcal{J}$, $\mathcal{F}$, and $\mathcal{J}\&\mathcal{F}$.

\noindent\textbf{Language-Guided Multi-Scale Dynamic Filtering.}
By comparing the third and fourth rows in Tab~\ref{tab:ablation_video}, we can find that the LMDF achieves the performance improvement by 2.5\%, 2.1\%, and 2.3\% in terms of $\mathcal{J}$, $\mathcal{F}$, and $\mathcal{J}\&\mathcal{F}$, respectively. This indicates that this module can accurately capture the temporal coherence between frames.\\
\emph{\textbf{Different Settings of LMDF.}}
We evaluate the design options about LMDF, and the results are shown in Tab.~\ref{tab:ablation_lmdf}.
By comparing the second (full LMDF) and third columns, it can be seen that learn multi-scale semantic context can produce more accurate segmentation mask.
The LMDF uses Eq.~\ref{dynamic_adp} to compute the position-adaptive guidance feature for dynamic filtering. While CMDy~\cite{wang2020context} and LGFS~\cite{hui2021collaborative} use max or average pooling to obtain the global guidance feature. Because their source codes are not available, we adjust the setting of LMDF to imitate the global language guidance for experimental comparison.
Specifically, we firstly use max pooling to deal with the multi-modal spatial-temporal feature $\mathrm{L}^c$ (generated by Eq.~\ref{dynamic_lan_guided}). And then, the pooled vector is used to weight the feature of current frame $\overline{\mathrm{V}}^c$ by element-wise multiplication. Thus, the global guidance information for each position is obtained.  Lastly, we generate dynamic kernels and filter the current frame by Eq.~\ref{dynamic}. The results (marked as MaxPool) show that the pooling operation indeed weakens the local perception required by the segmentation task.
In addition, LMDF interacts linguistic features with visual features in advance (Eq.~\ref{lag_ref}$\sim$Eq.~\ref{dynamic_lan_guided}), which makes the dynamic filters better align spatial-temporal grouping cues. The comparison between the second column and w/o pre-interaction also confirms this statement.
Unlike LGFS~\cite{hui2021collaborative}, which shares weight parameters across all spatial positions, LMDF learns the position-adaptive dynamic kernels. For comparison, we utilize $\mathrm{L}^c$ to generate the position-independent dynamic kernels.
Specifically, we conduct max pooling on $\mathrm{L}^c$ to obtain the global guidance vector, based on which the dynamic filters are generated the same way as in~\cite{hui2021collaborative}.
The results of the rightmost column in Tab.~\ref{tab:ablation_lmdf} show that the position-independent strategy leads to performance degradation.

\noindent\textbf{Qualitative cases.}
Fig.~\ref{fig:ablation_1} gives some representative examples to illustrate the benefits of our proposed module. We can find that the dual interleaved encoder can help the network accurately locate the referred object region, and the LMDF can utilize the temporal coherence to optimize the local details of the object. Thus, our model obtains a prediction mask that is closer to the ground-truth.

\section{Conclusion}
In this paper, we propose a vision-language interleaved dual encoder and a language-guided multi-scale dynamic filtering mechanism to address the referring video segmentation. Specifically, our method utilizes a language encoder, a visual encoder and a vision-language mutual guidance (VLMG) module to complete the progressive interweaving of different levels of multi-modal features between the two encoders. In addition, in order to fully depict the temporal coherence in the video, we further propose a language-guided multi-scale dynamic filtering (LMDF) module, which can learn adaptive spatial-temporal information with the guidance of language to promote the feature update of current frame.
Extensive experiments on four referring video segmentation datasets demonstrate that the proposed model significantly outperforms the state-of-the-art methods. The ablation study also verifies the effectiveness of the proposed modules.

{\small
\bibliographystyle{ieee_fullname}
\bibliography{egbib}

\begin{thebibliography}{10}\itemsep=-1pt

\bibitem{carreira2017quo}
Joao Carreira and Andrew Zisserman.
\newblock Quo vadis, action recognition? a new model and the kinetics dataset.
\newblock In {\em proceedings of the IEEE Conference on Computer Vision and
  Pattern Recognition}, pages 6299--6308, 2017.

\bibitem{dai2017deformable}
Jifeng Dai, Haozhi Qi, Yuwen Xiong, Yi Li, Guodong Zhang, Han Hu, and Yichen
  Wei.
\newblock Deformable convolutional networks.
\newblock In {\em Proceedings of the IEEE international conference on computer
  vision}, pages 764--773, 2017.

\bibitem{devlin2018bert}
Jacob Devlin, Ming-Wei Chang, Kenton Lee, and Kristina Toutanova.
\newblock Bert: Pre-training of deep bidirectional transformers for language
  understanding.
\newblock {\em arXiv preprint arXiv:1810.04805}, 2018.

\bibitem{ding2021vision}
Henghui Ding, Chang Liu, Suchen Wang, and Xudong Jiang.
\newblock Vision-language transformer and query generation for referring
  segmentation.
\newblock In {\em Proceedings of the IEEE/CVF International Conference on
  Computer Vision}, pages 16321--16330, 2021.

\bibitem{feng2021encoder}
Guang Feng, Zhiwei Hu, Lihe Zhang, and Huchuan Lu.
\newblock Encoder fusion network with co-attention embedding for referring
  image segmentation.
\newblock In {\em Proceedings of the IEEE/CVF Conference on Computer Vision and
  Pattern Recognition}, pages 15506--15515, 2021.

\bibitem{gavrilyuk2018actor}
Kirill Gavrilyuk, Amir Ghodrati, Zhenyang Li, and Cees~GM Snoek.
\newblock Actor and action video segmentation from a sentence.
\newblock In {\em Proceedings of the IEEE Conference on Computer Vision and
  Pattern Recognition}, pages 5958--5966, 2018.

\bibitem{hu2016segmentation}
Ronghang Hu, Marcus Rohrbach, and Trevor Darrell.
\newblock Segmentation from natural language expressions.
\newblock In {\em European Conference on Computer Vision}, pages 108--124.
  Springer, 2016.

\bibitem{hu2020bi}
Zhiwei Hu, Guang Feng, Jiayu Sun, Lihe Zhang, and Huchuan Lu.
\newblock Bi-directional relationship inferring network for referring image
  segmentation.
\newblock In {\em Proceedings of the IEEE/CVF Conference on Computer Vision and
  Pattern Recognition}, pages 4424--4433, 2020.

\bibitem{huang2020referring}
Shaofei Huang, Tianrui Hui, Si Liu, Guanbin Li, Yunchao Wei, Jizhong Han, Luoqi
  Liu, and Bo Li.
\newblock Referring image segmentation via cross-modal progressive
  comprehension.
\newblock In {\em Proceedings of the IEEE/CVF Conference on Computer Vision and
  Pattern Recognition}, pages 10488--10497, 2020.

\bibitem{hui2021collaborative}
Tianrui Hui, Shaofei Huang, Si Liu, Zihan Ding, Guanbin Li, Wenguan Wang,
  Jizhong Han, and Fei Wang.
\newblock Collaborative spatial-temporal modeling for language-queried video
  actor segmentation.
\newblock In {\em Proceedings of the IEEE/CVF Conference on Computer Vision and
  Pattern Recognition}, pages 4187--4196, 2021.

\bibitem{hui2020linguistic}
Tianrui Hui, Si Liu, Shaofei Huang, Guanbin Li, Sansi Yu, Faxi Zhang, and
  Jizhong Han.
\newblock Linguistic structure guided context modeling for referring image
  segmentation.
\newblock In {\em European Conference on Computer Vision}, pages 59--75.
  Springer, 2020.

\bibitem{jhuang2013towards}
Hueihan Jhuang, Juergen Gall, Silvia Zuffi, Cordelia Schmid, and Michael~J
  Black.
\newblock Towards understanding action recognition.
\newblock In {\em Proceedings of the IEEE international conference on computer
  vision}, pages 3192--3199, 2013.

\bibitem{khoreva2018video}
Anna Khoreva, Anna Rohrbach, and Bernt Schiele.
\newblock Video object segmentation with language referring expressions.
\newblock In {\em Asian Conference on Computer Vision}, pages 123--141.
  Springer, 2018.

\bibitem{li2018referring}
Ruiyu Li, Kaican Li, Yi-Chun Kuo, Michelle Shu, Xiaojuan Qi, Xiaoyong Shen, and
  Jiaya Jia.
\newblock Referring image segmentation via recurrent refinement networks.
\newblock In {\em Proceedings of the IEEE Conference on Computer Vision and
  Pattern Recognition}, pages 5745--5753, 2018.

\bibitem{li2017tracking}
Zhenyang Li, Ran Tao, Efstratios Gavves, Cees~GM Snoek, and Arnold~WM
  Smeulders.
\newblock Tracking by natural language specification.
\newblock In {\em Proceedings of the IEEE Conference on Computer Vision and
  Pattern Recognition}, pages 6495--6503, 2017.

\bibitem{liu2017recurrent}
Chenxi Liu, Zhe Lin, Xiaohui Shen, Jimei Yang, Xin Lu, and Alan Yuille.
\newblock Recurrent multimodal interaction for referring image segmentation.
\newblock In {\em Proceedings of the IEEE International Conference on Computer
  Vision}, pages 1271--1280, 2017.

\bibitem{liu2021cross}
Si Liu, Tianrui Hui, Shaofei Huang, Yunchao Wei, Bo Li, and Guanbin Li.
\newblock Cross-modal progressive comprehension for referring segmentation.
\newblock {\em IEEE Transactions on Pattern Analysis and Machine Intelligence},
  2021.

\bibitem{margffoy2018dynamic}
Edgar Margffoy-Tuay, Juan~C P{\'e}rez, Emilio Botero, and Pablo Arbel{\'a}ez.
\newblock Dynamic multimodal instance segmentation guided by natural language
  queries.
\newblock In {\em ECCV}, pages 630--645, 2018.

\bibitem{mcintosh2020visual}
Bruce McIntosh, Kevin Duarte, Yogesh~S Rawat, and Mubarak Shah.
\newblock Visual-textual capsule routing for text-based video segmentation.
\newblock In {\em Proceedings of the IEEE/CVF Conference on Computer Vision and
  Pattern Recognition}, pages 9942--9951, 2020.

\bibitem{ningpolar}
Ke Ning, Lingxi Xie, Fei Wu, and Qi Tian.
\newblock Polar relative positional encoding for video-language segmentation.
\newblock In {\em Proceedings of the Twenty-Ninth International Joint
  Conference on Artificial Intelligence, {IJCAI} 2020}, pages 948--954, 2020.

\bibitem{perazzi2016benchmark}
Federico Perazzi, Jordi Pont-Tuset, Brian McWilliams, Luc Van~Gool, Markus
  Gross, and Alexander Sorkine-Hornung.
\newblock A benchmark dataset and evaluation methodology for video object
  segmentation.
\newblock In {\em Proceedings of the IEEE conference on computer vision and
  pattern recognition}, pages 724--732, 2016.

\bibitem{seo2020urvos}
Seonguk Seo, Joon-Young Lee, and Bohyung Han.
\newblock Urvos: Unified referring video object segmentation network with a
  large-scale benchmark.
\newblock In {\em Proceedings of the European Conference on Computer Vision
  (ECCV)}, 2020.

\bibitem{shi2018key}
Hengcan Shi, Hongliang Li, Fanman Meng, and Qingbo Wu.
\newblock Key-word-aware network for referring expression image segmentation.
\newblock In {\em Proceedings of the European Conference on Computer Vision
  (ECCV)}, pages 38--54, 2018.

\bibitem{wang2020context}
Hao Wang, Cheng Deng, Fan Ma, and Yi Yang.
\newblock Context modulated dynamic networks for actor and action video
  segmentation with language queries.
\newblock In {\em Proceedings of the AAAI Conference on Artificial
  Intelligence}, volume~34, pages 12152--12159, 2020.

\bibitem{wang2019asymmetric}
Hao Wang, Cheng Deng, Junchi Yan, and Dacheng Tao.
\newblock Asymmetric cross-guided attention network for actor and action video
  segmentation from natural language query.
\newblock In {\em Proceedings of the IEEE/CVF International Conference on
  Computer Vision}, pages 3939--3948, 2019.

\bibitem{xu2015can}
Chenliang Xu, Shao-Hang Hsieh, Caiming Xiong, and Jason~J Corso.
\newblock Can humans fly? action understanding with multiple classes of actors.
\newblock In {\em Proceedings of the IEEE Conference on Computer Vision and
  Pattern Recognition}, pages 2264--2273, 2015.

\bibitem{ye2019cross}
Linwei Ye, Mrigank Rochan, Zhi Liu, and Yang Wang.
\newblock Cross-modal self-attention network for referring image segmentation.
\newblock In {\em Proceedings of the IEEE/CVF Conference on Computer Vision and
  Pattern Recognition}, pages 10502--10511, 2019.

\bibitem{ye2021referring}
Linwei Ye, Mrigank Rochan, Zhi Liu, Xiaoqin Zhang, and Yang Wang.
\newblock Referring segmentation in images and videos with cross-modal
  self-attention network.
\newblock {\em IEEE Transactions on Pattern Analysis and Machine Intelligence},
  2021.

\bibitem{yu2016modeling}
Licheng Yu, Patrick Poirson, Shan Yang, Alexander~C Berg, and Tamara~L Berg.
\newblock Modeling context in referring expressions.
\newblock In {\em ECCV}, pages 69--85. Springer, 2016.

\bibitem{zhang2020resnest}
Hang Zhang, Chongruo Wu, Zhongyue Zhang, Yi Zhu, Haibin Lin, Zhi Zhang, Yue
  Sun, Tong He, Jonas Mueller, R Manmatha, et~al.
\newblock Resnest: Split-attention networks.
\newblock {\em arXiv preprint arXiv:2004.08955}, 2020.

\end{thebibliography}
}

\end{document}